\documentclass[pmlr,twocolumn,10pt]{jmlr} 
\usepackage[T1]{fontenc}
\usepackage[utf8]{inputenc}

\mlhtrack{proceedings}

\newif\iffinal
\finaltrue  

\iffinal
    \ifmlhneedspmlr
      \jmlrvolume{346}
      \jmlryear{2026}
    \fi
    \ifmlhfindings \jmlrproceedings{}{ML4H 2026 - Findings Track}\fi
    \ifmlhdemo     \jmlrproceedings{}{ML4H 2026 - Demo Track}\fi
    \jmlrworkshop{Machine Learning for Health (ML4H) 2026}
\else
    \jmlrproceedings{}{Submitted to ML4H 2026: \mlhtrackname}
    \jmlrworkshop{Machine Learning for Health (ML4H) 2026}
\fi

\usepackage{booktabs}
\usepackage{siunitx}

\usepackage{dblfloatfix}

\usepackage[switch]{lineno}

\usepackage{listings}

\newcommand{\equal}[1]{{\hypersetup{linkcolor=black}\thanks{#1}}}
\newcommand{\hd}[1]{\begin{tabular}[b]{@{}c@{}}#1\end{tabular}}

\title[Distinguishing Extraction Failures
and Clinical Discordance with LLM Uncertainty on Radiology Reports]{Distinguishing Extraction Failures
and Clinical Discordance with LLM Uncertainty on Radiology Reports}

\author{Anonymous Author(s)}

\author{%
\Name{Jerjes Aguirre-Chavez}\equal{These authors contributed equally} \Email{jaguirrechavez@ucsd.edu}\\
\addr Hal{\i}c{\i}o\u{g}lu Data Science Institute, University of California, San Diego
\AND
\Name{Shant Malkasian}\footnotemark[1] \Email{smalkasian@health.ucsd.edu}\\
\addr Department of Radiology, University of California, San Diego
\AND
\Name{Roshun Sankaran} \Email{rsankaran@health.ucsd.edu}\\
\addr Department of Radiology, University of California, San Diego
\AND
\Name{Brendan Crabb} \Email{bcrabb@health.ucsd.edu}\\
\addr Department of Radiology, University of California, San Diego
\AND
\Name{Albert Hsiao} \Email{a3hsiao@health.ucsd.edu}\\
\addr Department of Radiology and Halıcıoğlu Data Science Institute, University of California, San Diego
}

\nolinenumbers 

\begin{document}

\maketitle

\ifmlhdemo\else
\begin{abstract}

Large language models (LLMs) are increasingly used to structure data from unstructured text, but extraction errors remain inevitable. We evaluated whether LLM uncertainty could distinguish extraction failure from true clinical discordance. In 5{,}225 vessels from 1{,}045 patients undergoing coronary CT angiography (CCTA) for suspected coronary artery disease (CAD), we evaluated token-level and self-reported LLM uncertainty during extraction of vessel-level stenosis severity. Extraction failure was defined as disagreement between LLM-extracted stenosis and expert annotations. Clinical discordance was defined as disagreement in obstructive CAD status between CCTA and subsequent invasive coronary angiography (ICA). Incremental discrimination was assessed after adjustment for stenosis severity, calcium burden, and coronary vessel.  Extraction failure was evaluated in 1{,}077 vessels from 253 patients with expert annotations, and clinical discordance in 1{,}641 vessels from 729 patients undergoing ICA within 90 days. Token-level Stenosis Margin showed the strongest association with extraction failure (adjusted OR, 4.15 [95\% CI, 2.57--6.69]; $P<0.001$) and clinical discordance (adjusted OR, 1.22 [95\% CI, 1.04--1.43]; $P=0.118$). Stenosis Margin significantly improved discrimination of extraction failure ($\Delta$AUC, +0.06 [95\% CI, +0.03 to +0.09]; $P=0.001$), but not clinical discordance ($\Delta$AUC, +0.01 [95\% CI, $-0.00$ to +0.02]; $P=0.245$). In a nested cohort of 138 vessels from 57 patients, 38 of 42 clinically discordant vessels (90\%) were extracted correctly. LLM uncertainty improved identification of extraction failure but not clinical discordance, suggesting that uncertainty may identify errors introduced during data extraction but may not generalize to discordance in underlying clinical data.

\end{abstract}
\fi

\ifmlhneedsstatements
\paragraph*{Data and Code Availability}
This study uses a retrospective single-institution cohort of coronary CT angiography and
invasive coronary angiography text reports dataset. Due to reports containing private health information markers, they are not released at this moment. To reproduce the analysis performed, we provide a GitHub repository:
\url{https://github.com/Jerjes0/llm_uncertainty_snippets} Analysis workflow is described in
\appendixref{apd:code}. The LLM prompts used are reproduced verbatim in
\appendixref{sec:ccta-prompt,sec:ica-prompt}.

\paragraph*{Institutional Review Board (IRB)}
This research was approved by our institutional review board with a waiver of informed
consent and was conducted in compliance with HIPAA (IRB \#201979).
\fi

\section{Introduction}
\label{sec:intro}

Structuring text into usable data for downstream analysis is paramount to medical and AI
research. In the field of medical imaging, AI has enabled large-scale analysis of image
and text datasets \citep{irvin2019chexpert}. Large Language Models (LLMs) have become the
pillar of agentic systems due to their capacity to integrate multiple data sources with
tools \citep{moor2023foundation,zhou2024medversa}. Although LLMs perform well in tasks
such as generation, summarization, and label extraction, existing evaluation is not
holistic. LLMs inevitably make mistakes \citep{kim2025hallucinations}. The reason behind
these mistakes may vary from lack of clarity in the task, inadequacy of the tools used, or
an inconsistency in the dataset itself \citep{stutz2025uncertain,chavoshi2025labelnoise}.
Distinguishing these failure sources is critical when extracted labels are used for
downstream scientific analysis.

Medical reports are highly variable across institutions, time periods, authors, and
language. However, the consequences of data variability are not evident. Large-scale label extraction from unstructured reports is often
a first step toward developing clinical tools
\citep{vanderloo2025structured,min2025cadrads}. However, errors in LLM label extraction may
further propagate in subsequent analyses, necessitating more detailed evaluation of
extraction performance, beyond typical measurements like accuracy or F1 scores
\citep{smit2020chexbert,irvin2019chexpert}. While these metrics quantify agreement with a
reference, further metrics are needed to help classify errors originating from extraction
failure or inherent issues with the source data.

A relevant aspect of LLM behavior that can be evaluated beyond aggregate accuracy is
uncertainty. Several families of approaches have been proposed to quantify uncertainty in
LLM outputs, including confidence elicitation, token-level probabilities, and consistency
across multiple generated responses. These measures provide signals of how reliably a
model arrives at a particular answer and can be evaluated by their ability to discriminate
correct from incorrect responses and by their calibration to observed accuracy
\citep{savage2025large}. Recent work has extended uncertainty estimation to structured
information extraction from clinical text, demonstrating that model self-derived
confidence can provide a quantitative measure of extraction reliability
\citep{schimmelpfennig2025certs}. More broadly, properties of clinical text that reflect
its predictability can also capture latent characteristics relevant to downstream decision
quality, suggesting that signals derived from language-model behavior may contain
information beyond the extracted content itself \citep{hsu2025machine}. However, whether
uncertainty provides useful information may depend on what aspect of an LLM-based pipeline
is being evaluated. In particular, uncertainty may indicate difficulty in extracting
information from a report without necessarily identifying whether the extracted
information is clinically concordant with subsequent observations.

A clinically important example of this challenge is the extraction of coronary artery
disease (CAD) severity from radiology reports, where measurements derived from coronary CT
angiography (CCTA) are routinely used to guide downstream diagnostic and treatment decisions. The
evaluation and risk stratification of patients with suspected CAD remains one of the
greatest challenges in modern medicine. Heart disease remains the leading cause of death
globally, primarily from CAD \citep{gbd2021usa}. Imaging plays a key role in the diagnosis
and treatment of CAD. Invasive coronary angiography (ICA) is the clinical reference
standard for diagnosing obstructive CAD (e.g.\ flow-limiting blockage within a coronary
artery), which guides decisions for escalating medical therapy or pursuing
revascularization \citep{lawton2022revascularization}. While ICA allows immediate
revascularization when obstructive CAD is present, most elective ICAs are unnecessary,
identifying obstructive CAD in less than half of referred patients \citep{patel2010yield}.
To this end, CCTA has emerged as a non-invasive ``gatekeeper'' to ICA. CCTA is now
guideline recommended for ruling-out obstructive CAD in patients with low-to-intermediate
risk of CAD, intending to reduce unnecessary downstream testing and improve clinical
outcomes \citep{zito2023strategies,gulati2021chestpain,vrints2024esc,douglas2015promise,%
williams2016scotheart}. While there is high certainty in a ``negative'' CCTA (sensitivity:
95\%; NPV: 98\%), the substantial uncertainty in a CCTA ``positive'' for obstructive CAD
(specificity: 79\%; PPV: 51\%) remains a major limitation
\citep{haase2019ipdmeta,serruys2021complete}. However, obstructive CAD is discordant in
nearly 25\% of patients between CCTA and ICA \citep{chang2019selective}, primarily due to
stenosis overestimation on CCTA. Discordances may arise from a combination of technical,
patient, and reader factors \citep{gulati2021chestpain,abdelrahman2020jacc,%
nieman2024standards,cury2022cadrads2}. The inherent technical constraints of CCTA, mainly
lower image and temporal resolution, limit its concordance with ICA
\citep{verpalen2026heartrate,fukui2025acquisition}. Patient factors such as coronary
calcification, high heart rate, and large body habitus introduce image artifacts which
compound with technical limitations, further degrading accurate measurement of stenosis
severity \citep{kruk2014calcium,andreini2021syntax3,kalisz2016artifacts}. Ultimately,
readers must approximate stenosis severity while accounting for these technical and
patient factors. Reader factors, particularly reader experience and habits, substantially
influence stenosis grading \citep{herzog2013experience,chow2021training,%
maroules2018interobserver,gouya2009discordance,lu2018corelab}. Less experienced readers
have worse diagnostic performance for identifying obstructive CAD on ICA
\citep{herzog2013experience,chow2021training}. Even among expert readers, interobserver
agreement of stenosis severity varies drastically (kappa: 0.77--0.85)
\citep{gouya2009discordance}. In a large trial, site readers overestimated stenosis nearly
twice as often as core lab readers---likely reflecting differences in experience and
reading frequency (daily for core lab vs.\ 1--3 times/week for site readers)
\citep{lu2018corelab}. The evaluation and treatment of CAD is based on the combination of
several factors, although stenosis severity remains one of the main drivers. Thus, LLMs
may be of particular interest in extracting CAD severity from CCTA and ICA reports.
Moreover, LLM uncertainty may provide an estimate of reader confidence in their image interpretation.

We present a retrospective analysis of CAD severity extraction from CCTA reports and
investigate whether LLM uncertainty can help identify and attribute failures in LLM-based
label extraction. We evaluate whether higher uncertainty is associated with extraction
errors, whether uncertainty provides incremental discrimination of CCTA--ICA discordance
beyond CCTA-reported CAD severity, and whether these represent distinct failure modes. We
hypothesize that LLM uncertainty is informative of extraction failure, provides additional
but task-dependent information about clinical discordance, and that uncertainty
informative of extraction errors does not necessarily transfer to clinical discordance.

\section{Methods and data}
\label{sec:methods}

\begin{figure*}[t]
\floatconts
  {fig:overview}
    {\caption{\textbf{LLM uncertainty against extraction failure and clinical
  discordance.} (A) Patients with suspected CAD undergo CCTA, with ICA as the clinical
  reference standard; CAD severity is categorized by percent stenosis. (B) An on-premises
  LLM platform extracts vessel-level stenosis severity from CCTA and ICA reports and
  derives the uncertainty measures. (C) Extraction failure is a mismatch against the
  expert CCTA annotation; (D) clinical discordance is a mismatch in obstructive CAD
  against ICA.}}
  {\includegraphics[width=1\linewidth]{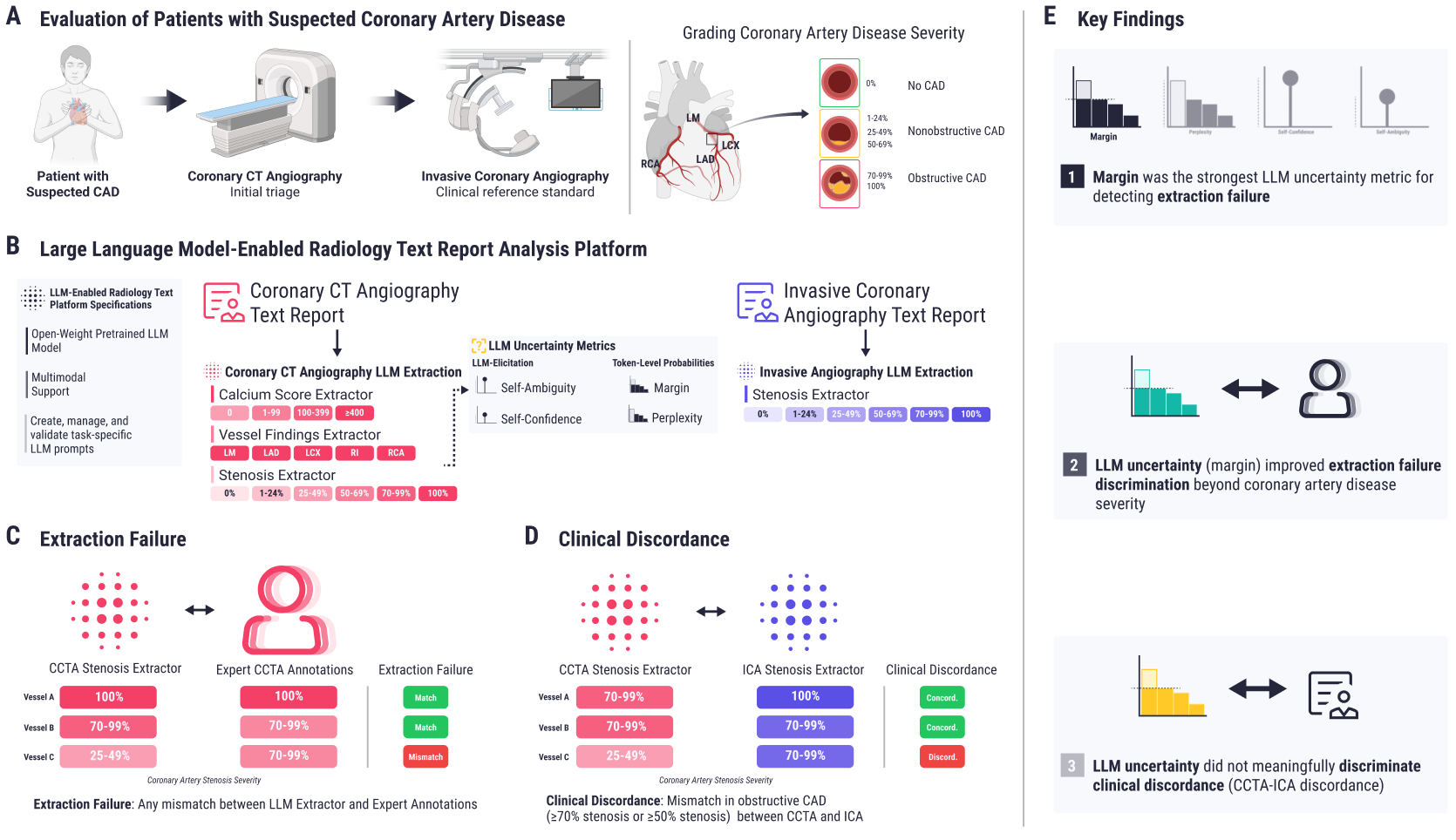}}
\end{figure*}

To distinguish extraction errors from clinical discordance, we use three measurements of
stenosis severity: stenosis severity extracted by the \textbf{CCTA LLM} from the CCTA report,
stenosis severity assigned by three \textbf{external radiologists} annotating the same text report,
and stenosis severity extracted by the \textbf{ICA LLM} from the subsequent ICA report. The
external radiologist reference provides an independent assessment of the information
contained in the CCTA report, while the ICA LLM provides an independent measurement of a
more reliable clinical reference.

With these three measurements of stenosis severity (CCTA LLM, CCTA reference annotations, and ICA LLM) we define two different failure modes to study: \textbf{Extraction Failure} and \textbf{Clinical Discordance}.
\textbf{Extraction Failure} is a disagreement between the stenosis severity dictated by
the CCTA LLM and the external radiologists. Similarly, \textbf{Clinical Discordance} is disagreement between the CCTA LLM and ICA LLM on vessel-specific obstructive CAD status.

Two families of uncertainty measures were used to characterize these failure modes: token-level statistics computed
over the generated response (margin, perplexity) and quantities elicited from the model
itself through prompting (self-confidence, self-ambiguity). In total, 10 unique uncertainty metrics were assessed, defined in \appendixref{apd:uncertainty}.  All metrics were rescaled to be directly proportional with LLM uncertainty for statistical analysis.

\subsection{Populations}

With IRB approval, HIPAA compliance, and a waiver of informed consent, we assembled a retrospective single-institution cohort of patients undergoing CCTA for evaluation of suspected CAD. Each patient contributed one index CCTA report. Patients with ICA preceding the index CCTA or prior coronary stent or coronary artery bypass graft were excluded.

Two reference populations were established. For the clinical-discordance analysis, patients were eligible if they underwent ICA within 90 days after CCTA; vessel-level inclusion additionally required both an extracted CCTA stenosis category and a quantified, vessel-matched ICA stenosis assessment. For the extraction-failure analysis, a subset of CCTA studies was independently reread by one of three external radiologists to establish reference stenosis annotations.

Four analysis populations were defined. P1 included patients with at least one reference source and all extracted vessels. P2 included vessels with both CCTA LLM extraction and external radiologist annotation and was used to evaluate Extraction Failure. P3 included vessels with both CCTA LLM extraction and vessel-matched ICA assessment and was used to evaluate Clinical Discordance. P4 included vessels with both reference sources and was used to directly compare the two failure modes.


\subsection{Extracting stenosis severity from CCTA and ICA text reports using LLMs}

Coronary artery stenosis severity was extracted from CCTA and ICA reports using open-weight LLMs.  For CCTA reports, vessel-level findings were separated using an LLM during pre-processing, for each of the main coronary artery vessels (left main, left anterior descending, left circumflex, right coronary, and ramus intermedius).  An LLM was then used to determine segment-level coronary stenosis severity, with a separate prompt applied to each of the five major coronary arteries. Segment-level results were subsequently aggregated to the vessel-level by assigning each vessel the maximum stenosis severity identified among its segments.  Stenosis severity was extracted on a six-point scale $\{0\%,\ 1\text{--}24\%,\ 25\text{--}49\%,\ 50\text{--}69\%,\ 70\text{--}99\%,\ 100\%\}$. For ICA reports, an LLM was applied directly to the findings section of each report, to extract vessel-level stenosis severity for the five main coronary arteries.  CCTA stenosis prompts are available in supplemental material (E.3 and E.4).

\subsection{LLM selection and comparison against traditional baseline}

Three open-weight LLMs were compared for stenosis severity extraction: Llama 3.3 70B \citep{grattafiori2024llama3}, MedGemma 27B \citep{sellergren2025medgemma}, and Qwen 3.5 9B \citep{qwen2025qwen35}. Stenosis extraction using regular expressions was also evaluated. Agreement with expert annotations was assessed using quadratic weighted Cohen's kappa and exact agreement stratified by stenosis severity category. To account for differences among the three tested models, the multi-model comparison used a separate, updated cohort export and was restricted to vessels with predictions available from all three models and a corresponding reference label.



\subsection{Extraction failure}

Extraction failure was defined as any disagreement between the LLM-extracted and expert-annotated stenosis severity, with concordance defined as exact agreement on the six-point stenosis severity scale.

\subsection{Clinical discordance}

Clinical discordance was defined as disagreement between the LLM-extracted CCTA and ICA measurements on the presence of obstructive CAD. Obstructive CAD was defined per vessel as $\geq 50\%$ stenosis in the left main coronary artery and $\geq 70\%$ stenosis in the left anterior descending, left circumflex, right coronary, and ramus intermedius arteries. Vessels were concordant when both measurements classified them as obstructive or both classified them as nonobstructive. Only vessels with an extracted CCTA stenosis category and a quantified vessel-matched ICA stenosis assessment were included. Vessels without a resolvable ICA stenosis measurement were excluded (\appendixref{sec:limitations}).

\subsection{Crossing extraction failure and clinical discordance}
\label{sec}

For failure-mode analyses, high versus low uncertainty was defined using the median Stenosis Margin in P2. Because smaller margins indicate greater uncertainty, values below this cutoff were classified as high uncertainty and values at or above it as low uncertainty. The cutoff was applied unchanged to P3 and P4. Each vessel was classified by extraction correctness and uncertainty; in P4, vessels were additionally classified by clinical concordance, allowing assessment of whether clinical discordance co-occurred with extraction failure.

To characterize differences between the two failure modes, we evaluated three report- or vessel-level characteristics: length of the vessel-specific report section provided to the LLM, the proportion of vessels with CCTA-reported stenosis $\geq70\%$, and coronary artery calcium score. These characteristics were evaluated separately for extraction failure in P2 and clinical discordance in P3, which comprised different vessel sets.

\subsection{Data aggregation}

All analyses were performed at the vessel level, with each patient contributing up to five observations corresponding to the main coronary arteries (\tableref{tab:populations}).

\subsection{Statistical analysis}

Confidence intervals and pairwise LLM comparisons were estimated using 2,000 patient-level bootstrap resamples to account for within-patient clustering. Pairwise $P$ values were Holm-corrected across the three LLM comparisons for each reference. For comparisons with regular expressions, Holm correction was applied across the three LLM-versus-regular-expression comparisons.

Logistic regression models were used to evaluate the ability of each uncertainty metric to discriminate extraction failure and clinical discordance. Generalized estimating equations with an exchangeable correlation structure were used to account for patient-level clustering. For each uncertainty metric, a univariable model was compared with a multivariable model adjusted for atherosclerotic plaque characteristics, including stenosis severity, coronary artery calcium score, and coronary vessel. Discrimination was assessed using the area under the receiver operating characteristic curve (AUC), with incremental improvement evaluated relative to a baseline model containing only atherosclerotic plaque characteristics. Differences in AUC were assessed using 2,000 bootstrap replicates. The Efron--Gong bootstrap procedure with 400 replicates was used to correct model performance for optimism. Model performance was additionally assessed using the Brier score and calibration curves based on deciles of predicted risk. The uncertainty metric with the largest AUC was carried forward for the main failure-mode analyses.

For the failure-mode analysis in P4, independence between extraction failure and clinical discordance was assessed using Fisher's exact test on the collapsed $2\times2$ correctness table. Because both classifications used the same uncertainty metric and cutoff, the uncertainty dimension was not included in this test. Given the limited P4 sample size ($n=138$ vessels), observed and expected co-occurrence of extraction failure and clinical discordance were also reported.

Report- and vessel-level characteristics were compared between correct and incorrect strata separately for extraction failure in P2 and clinical discordance in P3. Continuous variables were summarized as medians and categorical variables as percentages, with confidence intervals estimated using patient-level bootstrap resampling. Continuous variables were compared using the Mann--Whitney test and categorical variables using Fisher's exact test.

\section{Results}
\label{sec:results}

\subsection{Populations}

The source cohort included 7{,}847 CCTA studies from 7{,}847 patients. Patients with ICA preceding the index CCTA ($n=255$) or a prior coronary stent or coronary artery bypass graft ($n=302$) were excluded. Among the remaining patients, 847 underwent ICA within 90 days of CCTA, of whom 729 contributed vessel-matched data after requiring an extracted CCTA stenosis category and quantified ICA stenosis assessment. Separately, 270 CCTA studies were reread by one of three external radiologists. The ICA and external-radiologist reference populations overlapped in 72 patients, yielding 1{,}045 patients with at least one reference source before vessel-level eligibility requirements.

After vessel-level filtering, P1 included 5{,}225 vessels from 1{,}045 patients. The extraction-failure population (P2) included 1{,}077 vessels from 253 patients, and the clinical-discordance population (P3) included 1{,}641 vessels from 729 patients. Both reference measurements were available for the same vessel in 138 vessels from 57 patients (P4). Collectively, P2 and P3 included 925 unique patients.

\begin{table*}[t]
\floatconts
  {tab:populations}
  {\caption{\textbf{Primary analysis populations.} A row is one (patient-vessel) pair; each patient
  contributes one index CCTA and up to five vessels. All regression inference is clustered
  by patient.}}
  {\scriptsize\setlength{\tabcolsep}{4pt}\renewcommand{\arraystretch}{0.95}%
  \begin{tabular}{llrrrr}
    \toprule
    Abbreviation & Population & Analysis & Reference & Rows & Patients \\
    \midrule
    P1 & All extracted vessels & --- & --- & 5{,}225 & 1{,}045 \\
    P2 & CCTA $+$ external radiologist & Extraction Failure & External radiologist & 1{,}077 & 253 \\
    P3 & CCTA $+$ ICA & Clinical Discordance & ICA & 1{,}641 & 729 \\
    P4 & CCTA $+$ external radiologist $+$ ICA & Both & Both & 138 & 57 \\
    \bottomrule
  \end{tabular}}
\end{table*}


\subsection{LLM selection and comparison against traditional baseline}

For the multi-model comparison, 1{,}067 of the 1{,}077 P2 vessels from 253 patients had predictions available from all three LLMs and matching reference annotations. MedGemma and Qwen returned erroneous outputs in 10 vessels, which were excluded from the comparison. Against external radiologist annotations, quadratic weighted kappa ranged from 0.957 to 0.969 across LLMs. Against the ICA reference in the separate updated cohort export (1{,}724 shared vessels from 766 patients; P3), kappa ranged from 0.411 to 0.453 across LLMs, compared with 0.335 for regular expressions. Llama had the highest agreement point estimates against both references, although pairwise differences between LLMs against the external radiologist reference were not significant after Holm correction (\figureref{fig:llm-agreement}). In the separate regular-expression comparison, Llama achieved a kappa of 0.974 compared with 0.946 for regular expressions (Holm-adjusted $P=0.003$; \tableref{tab:llm-regex}). The accuracy of LLM stenosis extraction against reference annotations from 96 ICA reports is provided in \figureref{fig:ica-accuracy}.


\subsection{Extraction failure}

\textbf{Stenosis Margin}, the difference in probability between the two most likely tokens at the least confident position of the stenosis category, provided the strongest discrimination for extraction failure, and was selected as the metric for downstream analysis.

In P2 the CCTA LLM reproduced the external radiologists' category exactly on 967 of
1{,}077 vessels (89.8\%; quadratic weighted $\kappa=0.97$), leaving 110 extraction errors,
a rate of 10.2\% (\figureref{fig:agreement}A). The errors are small: 103 of 110 are
single-category, and only 7 vessels, 0.65\% of the population, differ by two or more
categories.

\begin{figure*}[!tb]
\floatconts
  {fig:agreement}
  {\caption{\textbf{Agreement between the CCTA LLM and each reference.}
  CCTA LLM stenosis category against (A) the assigned radiologist's category
  for the same report in P2 and (B) the ICA LLM category for the same vessel in P3.}}
  {\includegraphics[width=0.9\linewidth]{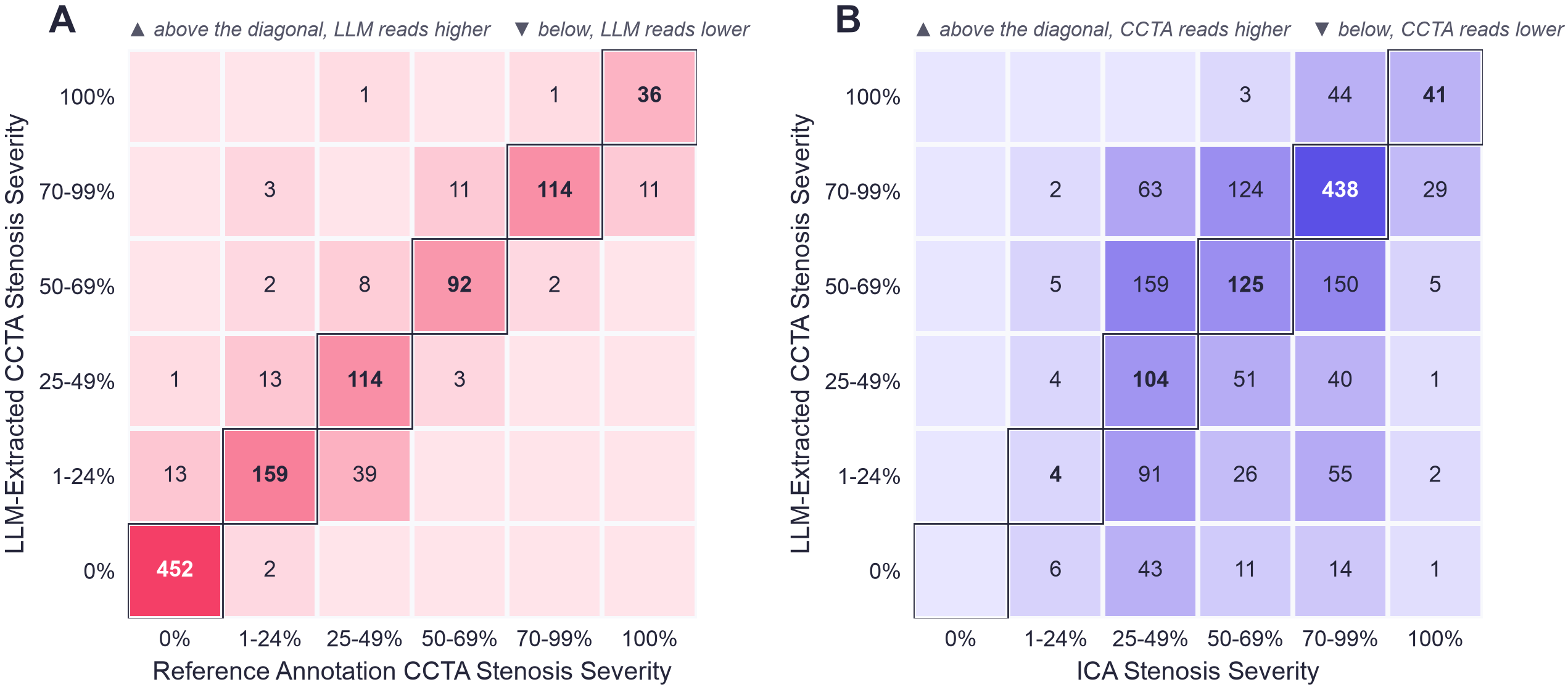}}
\end{figure*}

Uncertainty measures frequently associate with extraction error. After
adjustment for CAD phenotype, five survive Holm correction. Stenosis Margin leads at an
odds ratio of 4.15 (2.57--6.69) per standard deviation more uncertain, $P<0.001$, followed
by Ambiguity Margin at 1.66 (1.15--2.39), Self-Confidence at 1.62 (1.25--2.09),
Self-Ambiguity at 1.44 (1.17--1.77) and Stenosis Perplexity at 1.29 (1.12--1.48), all
Holm-adjusted $P\le0.027$.

Discrimination agrees with the associations. The CAD-phenotype baseline reaches an AUC of 0.805 (0.763--0.842), and adding Stenosis Margin raises it to 0.862 (0.828--0.892), a gain of $+0.057$ ($+0.029$, $+0.087$), bootstrap $P=0.001$ (\figureref{fig:roc-extraction}). The Brier score improves from 0.0788 to 0.0702 and the gain survives optimism correction, which moves the two models to 0.783 and 0.845. Stenosis Margin is therefore the selected uncertainty measure. It beats the runner-up by 0.037 AUC, outside that measure's own confidence interval, so the data resolve the choice. It is also a weak predictor on its own, at an AUC of 0.580, and gains its value only in combination with the CAD phenotype.

\subsection{Clinical discordance}

In P3 the CCTA LLM and the ICA LLM disagreed on obstructive status in 476 of 1{,}641
vessels (29.0\%). On the ordinal scale the two agreed exactly on 712 vessels (43.4\%) with
a quadratic weighted $\kappa$ of 0.45 (\figureref{fig:agreement}B).

Two measures show adjusted associations that survive Holm correction: Ambiguity Margin at
1.25 (1.10--1.43), $P=0.009$, and Self-Ambiguity at 1.19 (1.07--1.33), $P=0.018$. Both
belong to the self-reported ambiguity family, and they are the only two.

The baseline reaches an AUC of 0.618 (0.588--0.649), and adding the selected uncertainty measure moves it to 0.625 (0.595--0.655), a gain of $+0.007$ ($-0.004$, $+0.017$), $P=0.245$ (\figureref{fig:roc-clinical}). Across all measures the largest gain with an interval excluding zero is Self-Ambiguity at $+0.012$ ($+0.000$, $+0.024$), $P=0.049$, which is nominal and does not survive the near-collinearity of the ambiguity measures. 

\subsection{Failure distribution}

Splitting each population at the locked uncertainty cutpoints, gives the two four-state classifications in
\figureref{fig:distribution}. Under Extraction Failure the errors concentrate in the
uncertain half: 69 of 110 incorrect vessels (62.7\%) carry high uncertainty against 48.5\%
of correct ones, and only 41 vessels (3.8\% of P2) are confident and incorrect. Under
Clinical Discordance the same split is close to uninformative: high uncertainty accounts
for 43.7\% of discordant vessels and 46.3\% of concordant ones, and 268 vessels (16.3\% of
P3) are confident and discordant, four times the corresponding Extraction Failure cell.

\begin{figure*}[!tb]
\floatconts
  {fig:distribution}
  {\caption{\textbf{Failure distribution under each analysis.}
  LLM uncertainty crossed with (A) Extraction Failure in P2 and
  (B) Clinical Discordance in P3. Both panels use the median Stenosis Margin in P2 as the uncertainty cutoff, applied unchanged to P3.}}
  {\includegraphics[width=0.9\linewidth]{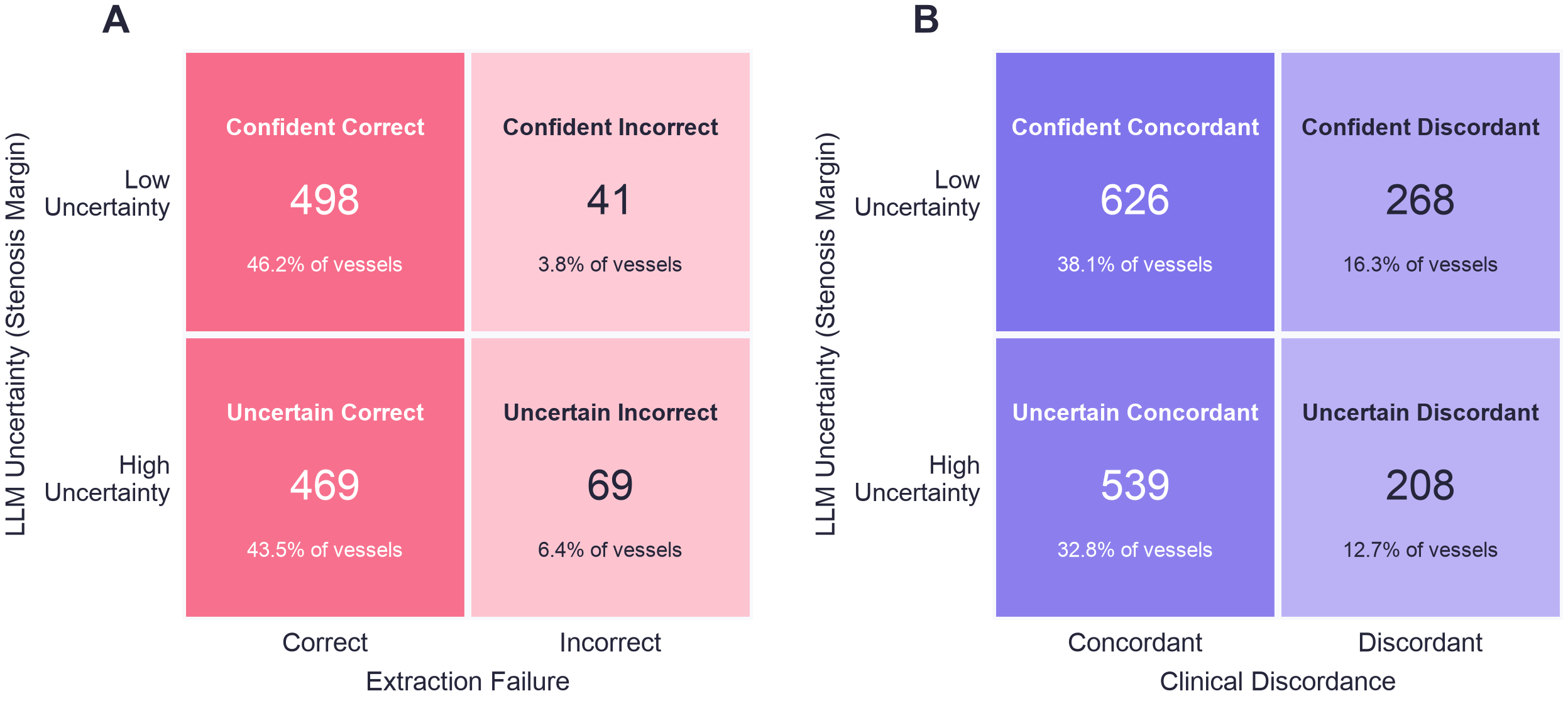}}
\end{figure*}

\subsection{Crossing extraction failure and clinical discordance}

There was little overlap between failure modes. In P4, 42 vessels are discordant and \textbf{38 of
them (90\%) were extracted correctly} (\figureref{fig:crossing}A), and only 4 of 16 extraction errors are also discordant.  The report- and vessel-level characteristics of the two failure modes also differed (\figureref{fig:crossing}B). Within P2, section length increased from a median of 187 to 244 characters ($P<0.001$), while the proportion of vessels with CCTA-reported stenosis $\geq70\%$ increased from 15.5\% to 24.5\% ($P=0.020$) and calcium score increased from a median of 0 to 64 ($P<0.001$). Within P3, the proportion of vessels with CCTA-reported stenosis $\geq70\%$ was higher in concordant than discordant vessels (47.7\% versus 39.5\%; $P=0.003$). Moreover, section length was similar between concordant and discordant vessels (276 versus 283 characters; $P=0.653$). 

\begin{figure*}[!tb]
\floatconts
  {fig:crossing}
  {\caption{\textbf{Extraction Failure against Clinical Discordance.}
  (A) Classification of P4 vessels by both outcomes; hatched cells are empty
  by construction. (B) Section length, the proportion of vessels with CCTA-reported stenosis $\geq70\%$, and calcium score by extraction correctness in P2 and clinical concordance in P3.}}
  {\includegraphics[width=0.9\linewidth]{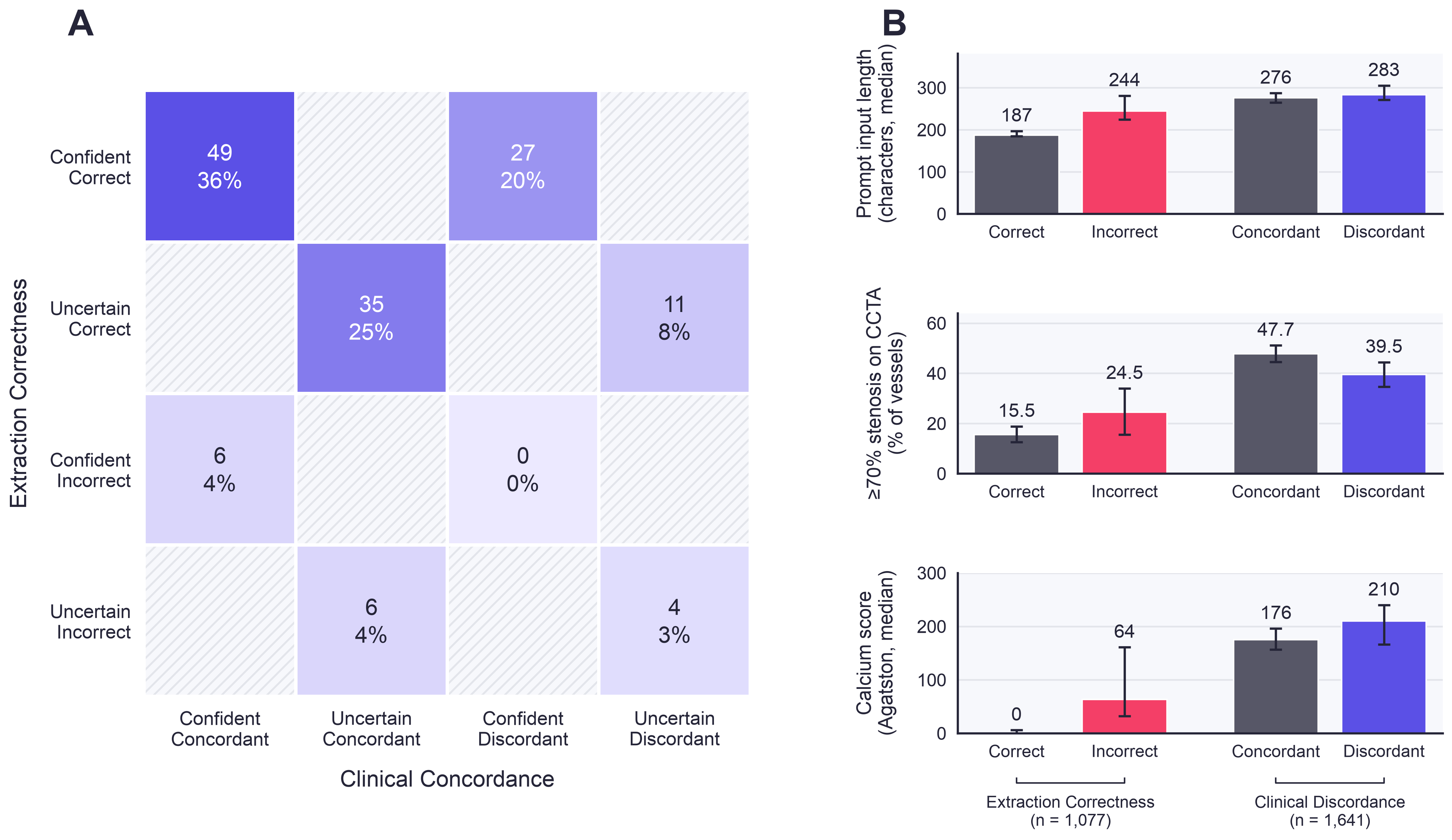}}
\end{figure*}

\section{Discussion}
\label{sec:discussion}

Our analyses demonstrate that LLM uncertainty can identify failures during LLM-based extraction of stenosis severity from radiology reports, but provides little additional discrimination of clinical discordance between stenosis severity assessed on CCTA and ICA. Higher LLM uncertainty was consistently associated with extraction failure, suggesting that uncertainty primarily reflects the reliability of the extraction process rather than agreement between the reported CCTA findings and the clinical reference standard.

Among the uncertainty measures evaluated, token-level measures, including margin and perplexity, showed the strongest associations with extraction failure. Stenosis Margin provided the greatest incremental discrimination beyond LLM-extracted CCTA stenosis severity and was therefore carried forward to subsequent analyses, whereas self-reported LLM uncertainty provided only minimal incremental improvement. These findings suggest that token-level uncertainty may provide a scalable mechanism for identifying cases in which LLM-based extraction is less reliable and may warrant additional review in large-scale clinical text extraction pipelines.

In contrast, although some uncertainty measures were associated with clinical discordance, their incremental discriminatory value was limited. Clinical discordance does not necessarily reflect difficulty extracting information from the CCTA report, but may instead arise from differences between CCTA and ICA, including technical, patient-related, and reader-related factors that are not captured by uncertainty generated during text extraction. Consistent with this interpretation, most clinically discordant vessels were extracted correctly, and clinical discordance demonstrated different associations with report and vessel characteristics than extraction failure. Uncertainty measures derived from LLM tasks specifically designed to assess the likelihood of CCTA–ICA discordance may provide greater discriminatory value and warrant evaluation in future studies.

Finally, comparison of the three LLMs demonstrated similar performance despite substantial differences in model size and pretraining. Although Llama 3.3 70B was selected for subsequent analyses, the comparable performance of the smaller, clinically pretrained MedGemma 27B model suggests that such models may offer a favorable tradeoff between performance and computational requirements. However, both Qwen 3.5 9B and MedGemma 27B performed similarly to the baseline regular-expression approach. Further fine-tuning using domain-specific datasets, particularly CCTA and ICA reports, may improve both extraction performance and the discriminatory value of LLM uncertainty measures.

Together, these findings indicate that the utility of LLM uncertainty is task dependent: uncertainty was informative for identifying extraction failure but did not meaningfully transfer to clinical discordance. LLM uncertainty should therefore not be considered a general measure of data reliability, and distinguishing extraction failures from discordance in the underlying clinical data may be important when assessing LLM-derived datasets for downstream analysis.


\clearpage

\bibliography{references}


\appendix

\onecolumn

\setcounter{figure}{0}
\renewcommand{\thefigure}{S\arabic{figure}}
\setcounter{table}{0}
\renewcommand{\thetable}{S\arabic{table}}

\section{Supplementary material}

\subsection{Uncertainty measures}
\label{apd:uncertainty}

The CCTA LLM answers one prompt per vessel and returns a stenosis category for each
segment. We record two token-level statistics over that response: the minimum
token margin, the gap between the two most probable tokens at the least confident position, and
the perplexity of the generated sequence. A vessel takes the maximum severity among its segments,
so we pool both statistics over the segments tied at that maximum: the minimum margin and the
maximum perplexity among the segments that actually determine the vessel's label. We call these
\textbf{Stenosis Margin} and \textbf{Stenosis Perplexity}.

In the same call we ask the model to rate its own reading on two scales from 0 to 1, one for how
confident it is and one for how ambiguous the report was. These are \textbf{Self-Confidence} and
\textbf{Self-Ambiguity}, and they are reported per vessel rather than per segment. The prompts that elicit them are reproduced in Sections~\ref{sec:ccta-prompt} and~\ref{sec:ica-prompt}.

Margin and perplexity can be computed over any span of generated tokens, not only the stenosis
category. We therefore compute them over three further spans: the confidence and ambiguity fields
of the same response, and the separate call that extracts the findings section the stenosis prompt
reads. This gives six additional measures --- Findings, Confidence and Ambiguity Margin and
Perplexity --- and ten in total. The confidence and ambiguity pairs ask whether how fluently the
model states a score carries information the score itself does not. Agreement among the ten
measures is reported in Figure~\ref{fig:s5}.

Higher values mean more uncertainty for every measure, so we negate margin and self-confidence,
take perplexity on the log scale, and standardize each measure within the population being
modelled. The two Findings measures are available for 46\% of P2, concentrated in the left
anterior descending artery, and we analyze them complete-case rather than let them reduce the
population. Figure~\ref{fig:s1} marks the model that could not be fitted as a result.

\section{Technical details}

All extraction ran locally on institutional hardware; no report text left the secure
environment and no hosted model API was used. Every extraction used the same open-weight
model, \texttt{Llama-3.3-70B-Instruct} (\texttt{unsloth} release) served through vLLM,
few-shot with no fine-tuning, differing only in its registered prompt and output
schema. Inference was distributed across four GPUs on two internal servers: two
NVIDIA RTX A6000 (48\,GB) on the first, and one RTX A6000 (48\,GB) plus one RTX A4500
(20\,GB) on the second.

\section{Code}
\label{apd:code}

The analysis is available at
\url{https://github.com/Jerjes0/llm_uncertainty_snippets}. It holds the eight
numbered stages that build the cohorts, fit the regressions, compute discrimination and
calibration, and lock the uncertainty measure and its cutoff, together with two further
stages that assemble the manuscript figures. Nothing in it imports from the extraction
platform. The pipeline starts from a cohort export and never touches a model, a prompt
registry or a database, so what is released is the statistical analysis rather than the
clinical system that produced the extractions.

The data are not included, so the code documents the analysis rather than permitting
end-to-end re-execution. The repository states the column contract a cohort export must
satisfy, down to the severity vocabulary and the treatment of an unquantified angiographic
vessel, and carries a script that checks a substitute cohort against that contract from
the CSV headers alone.

\section{Limitations}
\label{sec:limitations}

This study has several limitations. A vessel enters P3 only when evaluated on ICA, which removes 1{,}729 of 3{,}370 vessels and does so non-randomly, since
a healthy vessel is the one most likely to go unquantified. P3 is therefore enriched for
disease, and the 29.0\% discordance rate and the baseline AUC of 0.618 describe that
restricted population rather than the full ICA-eligible cohort. The 270 rereads were
selected manually rather than at random, and each report was read by one of three
radiologists without adjudication, so we cannot separate an extraction error from a
disagreement between readers. The two failure modes are also measured on different
populations under different endpoints, an exact six-point match against a binary
obstructive threshold, so the contrast in incremental discrimination reflects the two
tasks as a whole rather than the nature of the failure alone. P4 holds 138 vessels,
enough to exclude a strong association between the two failure modes but not a weak one.
Finally, this is a single institution and a single open-weight model, and Stenosis Margin
requires access to token probabilities that a hosted model would not expose.

\section{Extraction details and supplementary results}

\subsection{Abbreviations}

\begin{table}[t]
\floatconts
  {tab:abbrev}
  {\caption{Abbreviations used throughout the paper.}}
  {\footnotesize
  \begin{tabular}{ll}
    \toprule
    Abbreviation & Full name \\
    \midrule
    \multicolumn{2}{l}{\emph{Coronary vessels}} \\
    LM   & Left main coronary artery \\
    LAD  & Left anterior descending artery \\
    LCX  & Left circumflex artery \\
    RCA  & Right coronary artery \\
    RI   & Ramus intermedius \\
    \addlinespace
    \multicolumn{2}{l}{\emph{Coronary segments}} \\
    PLAD / MLAD / DLAD  & Proximal / mid / distal left anterior descending artery \\
    D1 / D2             & First / second diagonal branch \\
    PCX / MCX           & Proximal / mid left circumflex artery \\
    OM1 / OM2           & First / second obtuse marginal branch \\
    LPLB / LPDA         & Left posterolateral branch / left posterior descending artery \\
    PRCA / MRCA / DRCA  & Proximal / mid / distal right coronary artery \\
    RPLB / RPDA         & Right posterolateral branch / right posterior descending artery \\
    \addlinespace
    \multicolumn{2}{l}{\emph{Imaging studies and measurements}} \\
    CCTA     & Coronary computed tomography angiography \\
    ICA      & Invasive coronary angiography \\
    CACS     & Coronary artery calcium score (Agatston units) \\
    CAD-RADS & Coronary Artery Disease Reporting and Data System \\
    NR       & Not reported (stenosis severity not stated for that segment) \\
    \addlinespace
    \multicolumn{2}{l}{\emph{Methods and statistics}} \\
    LLM   & Large language model \\
    AUC   & Area under the receiver operating characteristic curve \\
    OR    & Odds ratio \\
    IRB   & Institutional Review Board \\
    HIPAA & Health Insurance Portability and Accountability Act \\
    \bottomrule
  \end{tabular}}
\end{table}

\subsection{Segment-to-vessel aggregation}

The CCTA LLM returns a stenosis category per coronary segment. A vessel takes the maximum
severity among the segments feeding it, listed in Table~\ref{tab:s0}. The ramus intermedius is
kept as a vessel of its own rather than folded into the left anterior descending artery, giving
five vessels per patient. Each vessel is read by one prompt, and the angiographic vessel it is
compared against carries the same name.

\begin{table}[t]
\floatconts
  {tab:s0}
  {\caption{Segment-to-vessel aggregation. Eighteen segments collapse to five vessels.}}
  {\small
  \begin{tabular}{lll}
    \toprule
    Vessel & Segments & ICA vessel \\
    \midrule
    Left main (LM) & LM & LM \\
    Left anterior descending (LAD) & proximal, mid, distal LAD; D1; D2 & LAD \\
    Ramus intermedius (RI) & RI & RI \\
    Left circumflex (LCX) & proximal, mid LCX; OM1; OM2; L-PLB; L-PDA & LCX \\
    Right coronary (RCA) & proximal, mid, distal RCA; R-PLB; R-PDA & RCA \\
    \bottomrule
  \end{tabular}}
\end{table}

\subsection{Prompts}
\label{sec:prompts}
Each prompt states the segments to report, the six-point severity scale, the mapping from
descriptive phrasing to a category, and a set of worked examples drawn from real reports. Every
prompt also elicits the two self-reported scores. The angiographic prompt returns a percentage or
the literal string \texttt{none}, which is the ambiguity discussed in the Clinical
discordance section of the main text.


\subsection{CCTA LLM prompt}
\label{sec:ccta-prompt}

The prompt below is reproduced verbatim as registered in the prompt registry
(v4.0.0). It is the prompt for the LAD territory; the LM, LCX and RCA prompts follow
the same structure, differing only in the segment list and in territory-specific
worked examples. One prompt is issued per vessel. The worked examples are
de-identified report excerpts and contain no protected health information.

\begin{lstlisting}[basicstyle=\ttfamily\scriptsize, breaklines=true,
                   columns=fullflexible, frame=single, framesep=4pt, xleftmargin=2pt]
INSTRUCTIONS
Your task is to extract the following information from a coronary CT angiography text report:

- Use the following terms to report the maximum stenosis severity for each coronary segment (STENOSIS RANGE = STENOSIS SEVERITY INTERPRETATION):
 0% = ABSENT
 1-24% = MINIMAL
 25-49% = MILD
 50-69% = MODERATE
 70-99% = SEVERE
 100% = TOTAL OCCLUSION
 NR = NOT REPORTED

- Report stenosis severity for the following coronary segments (SEGMENT ABBREVIATION = SEGMENT NAME):
 PLAD = PROXIMAL LEFT ANTERIOR DESCENDING
 MLAD = MID LEFT ANTERIOR DESCENDING
 DLAD = DISTAL LEFT ANTERIOR DESCENDING
 D1 = FIRST DIAGONAL
 D2 = SECOND DIAGONAL

- Do not use vessel "caliber" size to determine stenosis severity

- If a range of stenosis severity is provided, report the lower boundary stenosis severity; here are some examples:
    - "up to mild stenosis" should have a stenosis severity of "1-24%"
    - "approaching moderate stenosis" should have a stenosis severity of "25-49%"
    - "mild to moderate stenosis" should have a stenosis severity of "25-49%"
    - "moderate to severe stenosis" should have a stenosis severity of "50-69%"

- The following statements are equivalent to 1-24%:
  - "luminal irregularities"
  - "mild plaque without significant stenosis"
  - any reference to plaque without an explicit degree of stenosis stated

- The following statements are equivalent to 70-99%:
  - "high-grade stenosis"

CONFIDENCE
- Report how confident you are that the stenosis category you selected is the correct interpretation of the report text according to the extraction instructions. Use a scalar from 0.0 to 1.0, where 0.0 indicates no confidence and 1.0 indicates complete confidence. Use the full range of values.

AMBIGUITY
- Report the degree to which the report text permits multiple reasonable stenosis-category interpretations for the segment, independent of your confidence in the category you ultimately selected.
- 0.0 = completely unambiguous; the relevant finding is explicit and supports essentially one interpretation
- 1.0 = highly ambiguous; the text is vague, conflicting, equivocal, difficult to localize, or supports multiple reasonable interpretations

**Confidence and source ambiguity are distinct. A report may contain an ambiguous or equivocal stenosis description that you nevertheless extract with high confidence according to the rules above. Conversely, an apparently definitive statement may still be difficult to attribute to the correct coronary segment. Rate each independently.**

EXAMPLES

[Thirteen worked examples, each a de-identified LAD excerpt followed by its
EXPECTED RESPONSE block giving PLAD / MLAD / DLAD / D1 / D2 categories and
placeholders for CONFIDENCE and AMBIGUITY.]

OUTPUT FORMAT
Extract the worst stenosis in each coronary segment using YAML structuring, following this format and keys:
 PLAD: 0%, or 1-24%, or 25-49%, or 50-69%, or 70-99%, or 100%, or NR
 MLAD: 0%, or 1-24%, or 25-49%, or 50-69%, or 70-99%, or 100%, or NR
 DLAD: 0%, or 1-24%, or 25-49%, or 50-69%, or 70-99%, or 100%, or NR
 D1: 0%, or 1-24%, or 25-49%, or 50-69%, or 70-99%, or 100%, or NR
 D2: 0%, or 1-24%, or 25-49%, or 50-69%, or 70-99%, or 100%, or NR
 CONFIDENCE: A measure ranging from 0.0 to 1.0 describing the level of confidence that the stenosis measurements accurately reflect what is in the report
 AMBIGUITY: A measure ranging from 0.0 to 1.0 describing the level of ambiguity of the stenosis measurements described in the report

Only output the YAML serialized result.  DO NOT INCLUDE ANY ADDITIONAL INFORMATION OR EXPLANATIONS.
\end{lstlisting}

\subsection{ICA LLM prompt}
\label{sec:ica-prompt}

The prompt below is reproduced verbatim as registered in the prompt registry
(\texttt{llm\_parse\_ica\_stenosis}, v1.0.0). The five worked examples are
de-identified report excerpts and contain no protected health information; they are
part of the prompt itself and were sent with every request.

\begin{lstlisting}[basicstyle=\ttfamily\scriptsize, breaklines=true,
                   columns=fullflexible, frame=single, framesep=4pt, xleftmargin=2pt]
# INSTRUCTIONS

You are the world's greatest interventional cardiologist.  Extract information about diameter stenosis results from an invasive coronary angiography text report.

Extract the following information:
- lm:  maximum diameter stenosis in the left main coronary artery (only includes the left main)
- lad:  maximum diameter stenosis in the left anterior descending coronary artery (includes the following segments:  proximal LAD, mid LAD, distal LAD, diagonal branches)
- lcx:  maximum diameter stenosis in the left circumflex coronary artery (includes the following segments:  proximal CX, mid CX, left PLB/PLV, left PDA, obtuse marginal branches)
- rca:  maximum diameter stenosis in the right coronary artery (includes the following segments:  proximal RCA, mid RCA, distal RCA, right PLB/PLV, right PDA)
- ri:  maximum diameter stenosis in the ramus intermedius (includes only the ramus intermedius)

## SPECIAL CASES
- If multiple diameter stenoses are reported in a coronary territory, extract the maximum diameter stenosis measurement

# EXAMPLES

[Five worked examples, each a de-identified ANGIOGRAPHIC FINDINGS excerpt followed by
its EXPECTED RESULTS block. Reproduced in full in the supplementary prompt listing.]

# OUTPUT FORMAT

Extract maximum diameter stenosis measurements of the LM, LCX, RCA, and RI territories for the following text report.  Extract continuous diameter stenosis measurements in YAML using the following structure; report "none" if no diameter stenosis measurement is available for that territory.  *DO NOT INCLUDE ANY ADDITIONAL INFORMATION BEYOND THIS*

- lm: none or maximum diameter stenosis
- lad: none or maximum diameter stenosis
- lcx: none or maximum diameter stenosis
- rca: none or maximum diameter stenosis
- ri: none or maximum diameter stenosis
\end{lstlisting}

\clearpage
\subsection{Supplementary figures}

\begin{figure}[t]
\floatconts
  {fig:s1}
  {\caption{\textbf{All uncertainty measures against Extraction Failure (P2).} (A) Extraction
  error rate by quintile of each measure, lightest to darkest for least to most uncertain. (B) AUC
  for each measure alone, for the CAD-phenotype baseline, and for the baseline with that measure
  added, with the change against the baseline at right. Findings Margin is not estimable because
  the findings section is recovered for a minority of vessels.}}
  {\includegraphics[width=\linewidth]{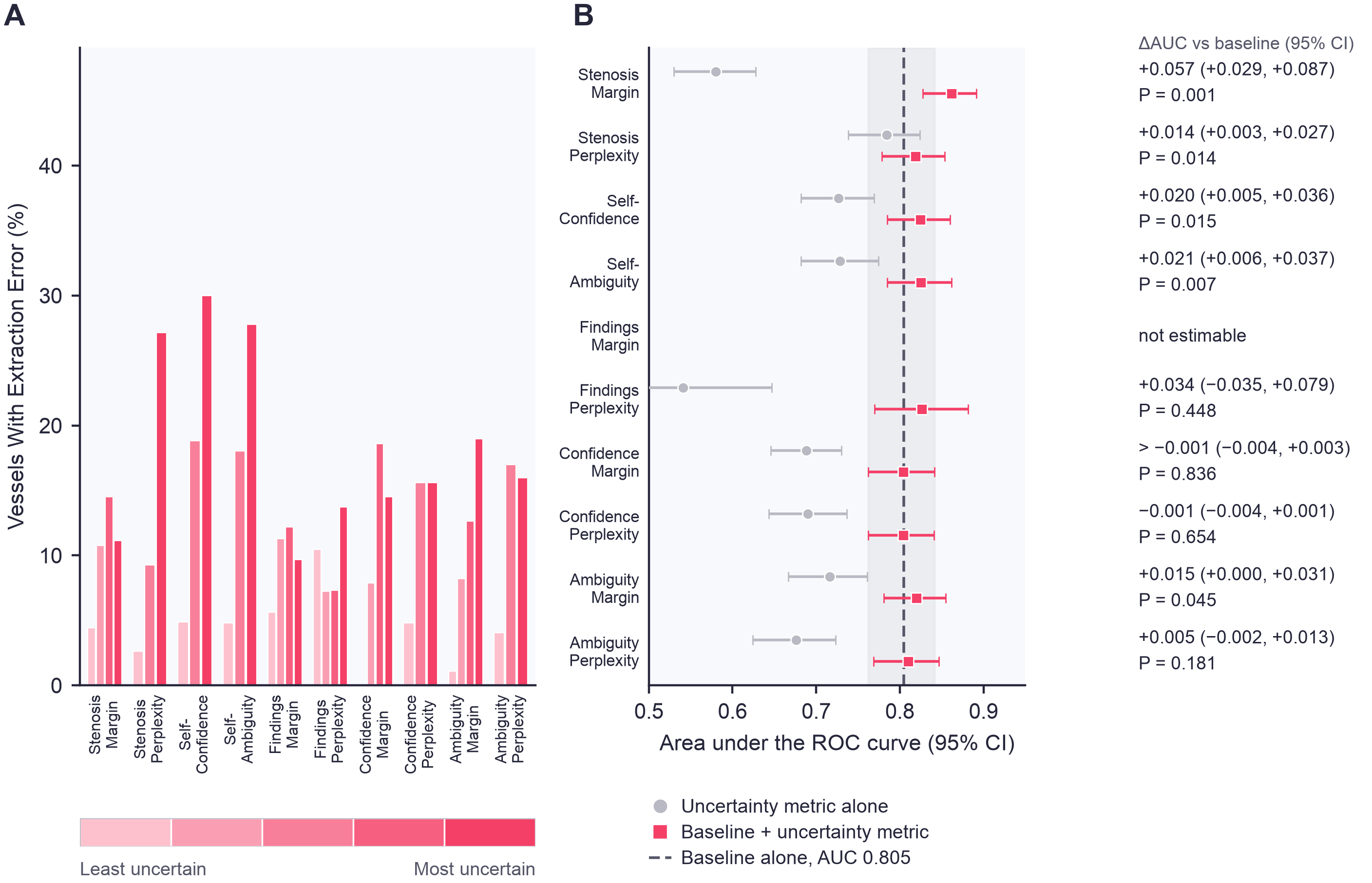}}
\end{figure}

\begin{figure}[t]
\floatconts
  {fig:s2}
  {\caption{\textbf{All uncertainty measures against Clinical Discordance (P3).} Panels as in
  Figure~\ref{fig:s1}. The measure and cutoff selected in P2 are carried over without
  recalculation.}}
  {\includegraphics[width=\linewidth]{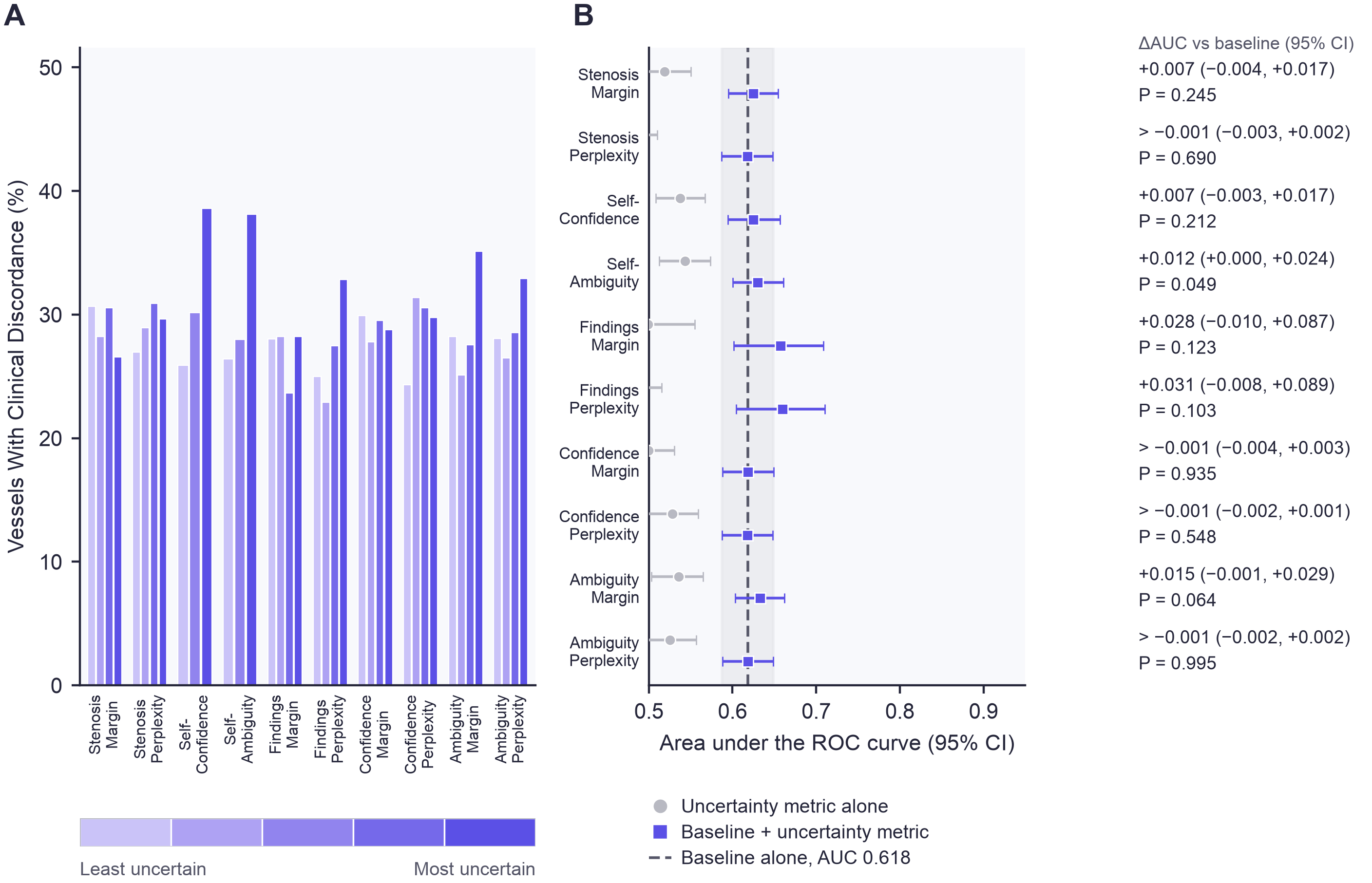}}
\end{figure}

\begin{figure}[t]
\floatconts
  {fig:s3}
  {\caption{\textbf{Calibration, Extraction Failure (P2).} Equal-count bins of predicted risk
  with Wilson intervals; the dashed line is perfect calibration. Each panel gives the Brier score
  and the value after optimism correction.}}
  {\includegraphics[width=\linewidth]{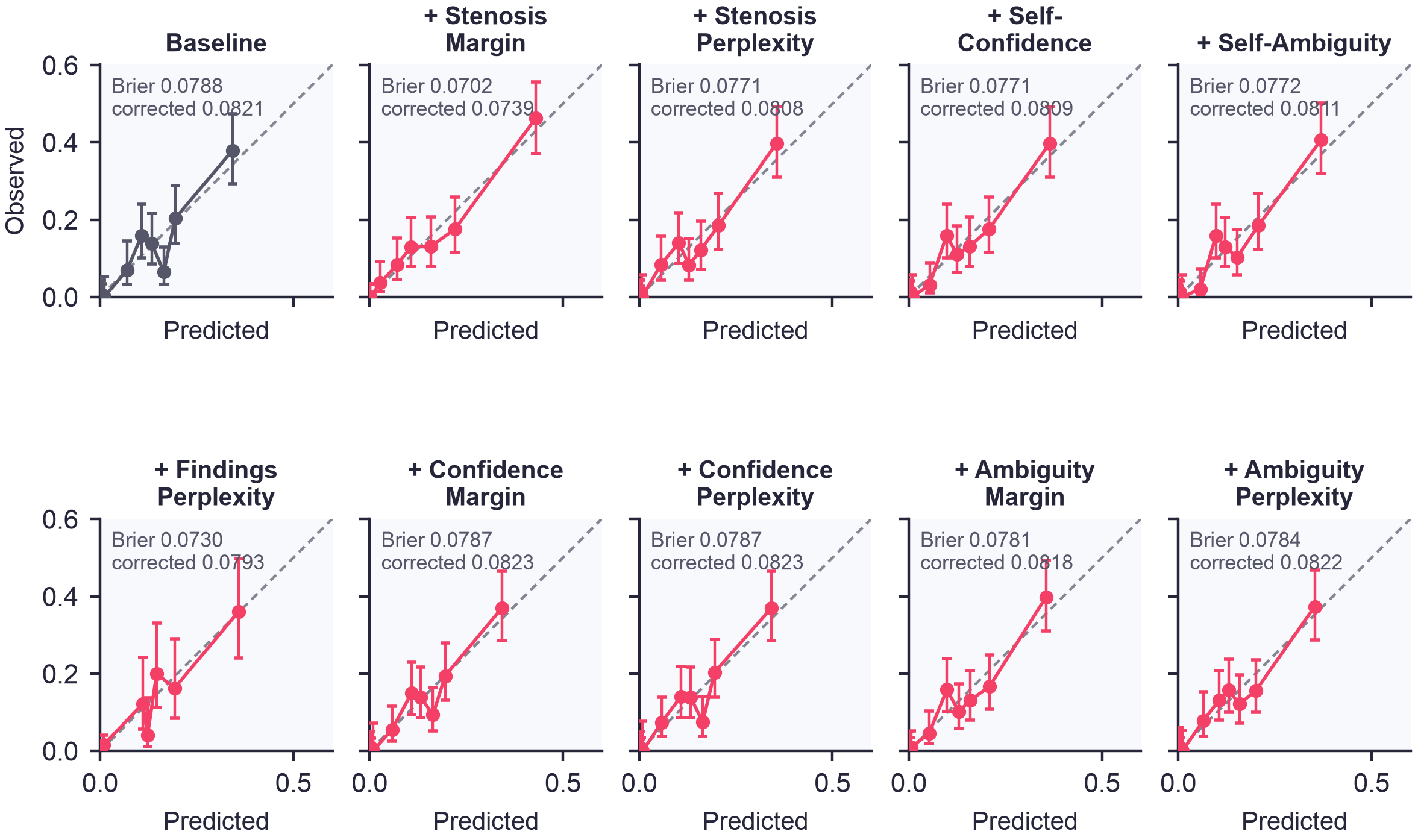}}
\end{figure}

\begin{figure}[t]
\floatconts
  {fig:s4}
  {\caption{\textbf{Calibration, Clinical Discordance (P3).} Panels as in Figure~\ref{fig:s3}.}}
  {\includegraphics[width=\linewidth]{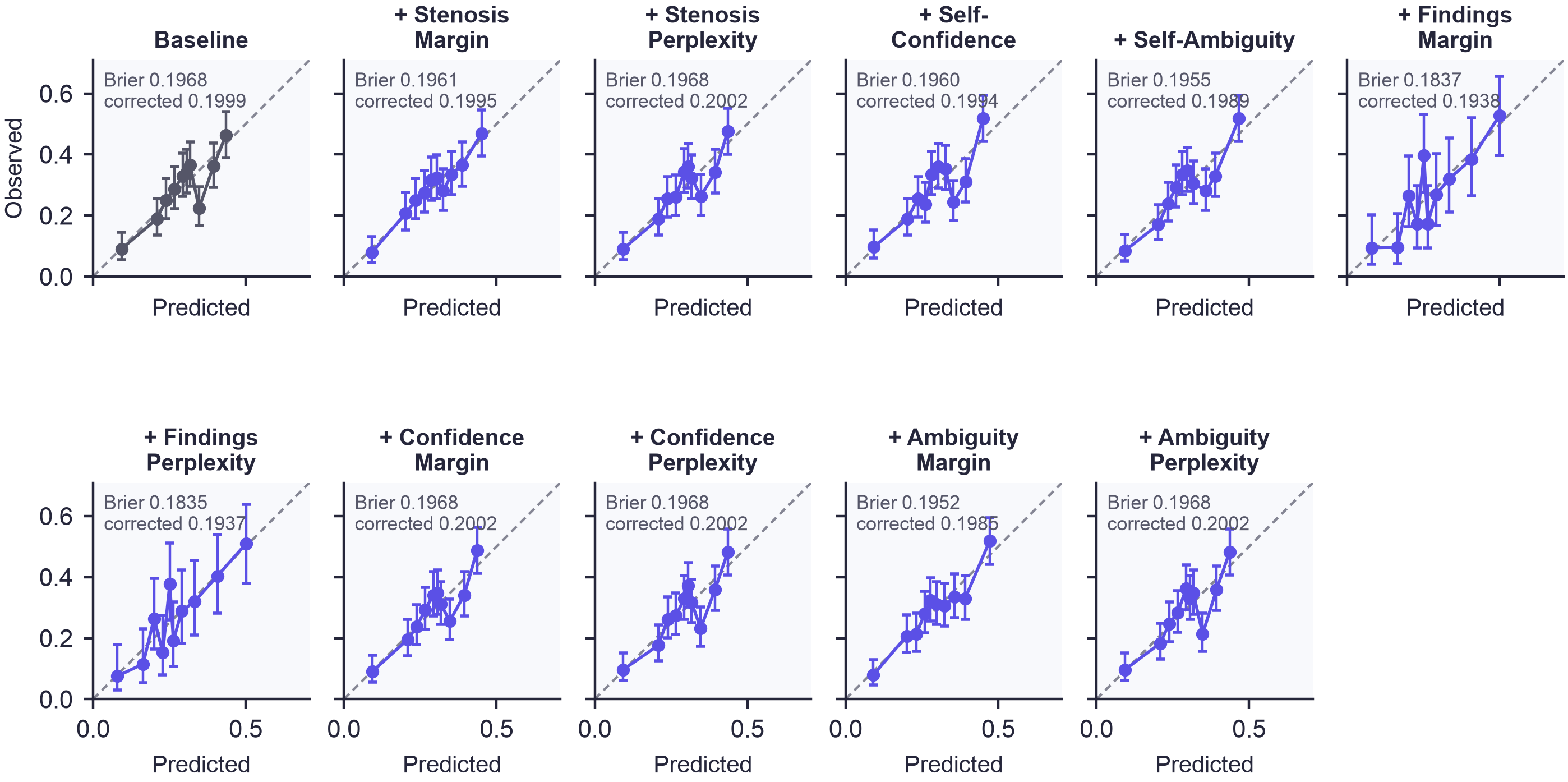}}
\end{figure}

\begin{figure}[t]
\floatconts
  {fig:s5}
  {\caption{\textbf{Agreement among the token-level measures.} Pairwise scatter with ordinary
  least-squares fits, Pearson $r$ and Spearman $\rho$, for the four margin measures (A) and the
  four perplexity measures (B), in P2.}}
  {\includegraphics[width=\linewidth]{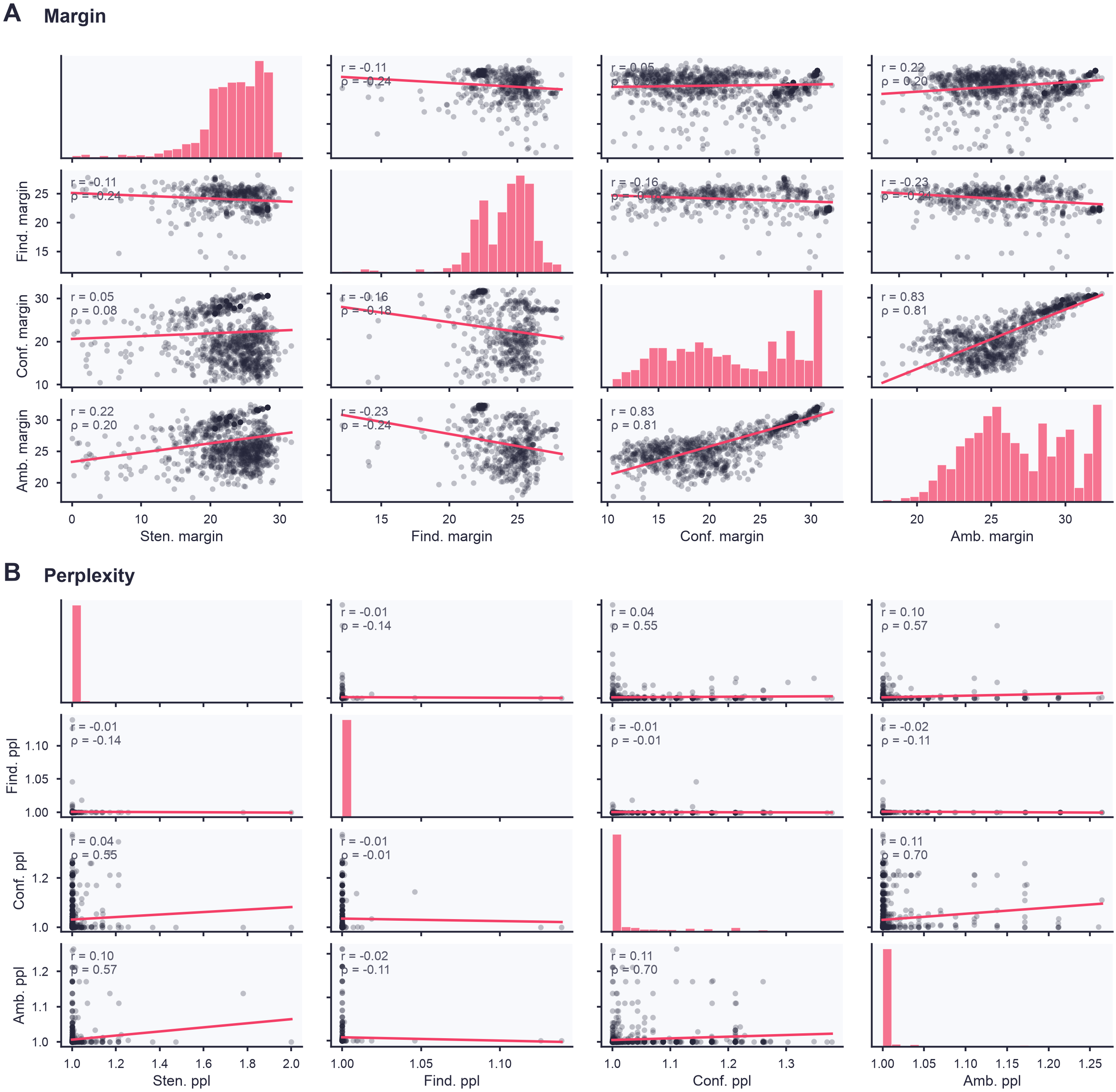}}
\end{figure}

\begin{figure}[t]
\floatconts
  {fig:ica-accuracy}
  {\caption{\textbf{ICA extraction accuracy.} ICA LLM stenosis category against the reference
  annotation for the same vessel, over 480 vessels from 96 ICA reports, diagonal outlined. Exact
  agreement on 469 vessels (97.7\%), quadratic weighted $\kappa=0.98$. Eight of the eleven
  disagreements sit above the diagonal, so the ICA LLM read higher than the reference.}}
  {\includegraphics[width=0.6\linewidth]{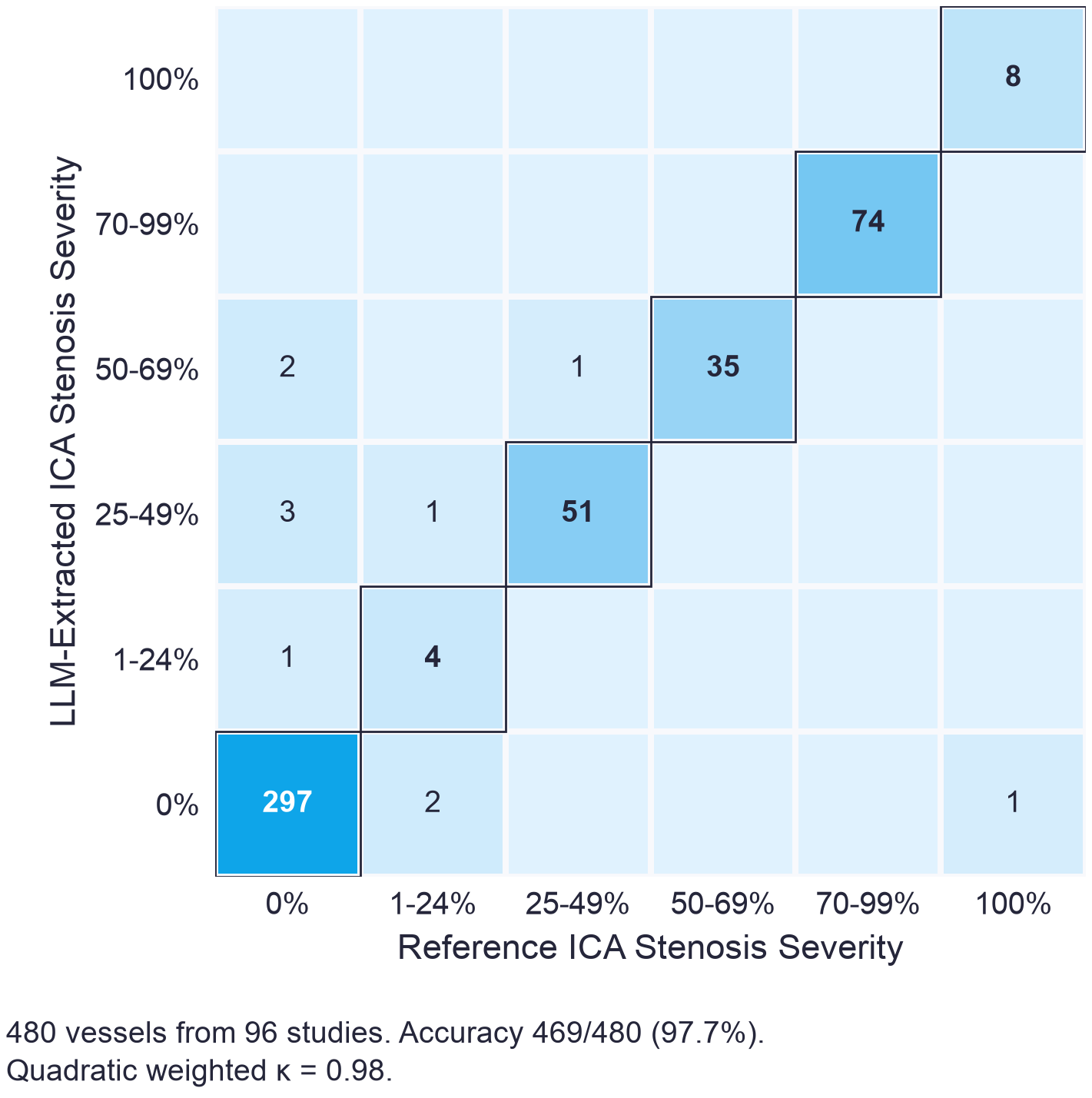}}
\end{figure}

\begin{figure}[t]
\floatconts
  {fig:llm-agreement}
  {\caption{\textbf{Agreement across LLMs and reference measurements.}
  Pairwise quadratic weighted $\kappa$ among the three LLMs and
  (A) assigned radiologist annotations (1{,}067 shared vessels; 253 patients) or
  (B) the LLM-extracted ICA reference (1{,}724 shared vessels; 766 patients).
  Error bars show patient-bootstrap 95\% confidence intervals.}}
  {\includegraphics[width=\linewidth]{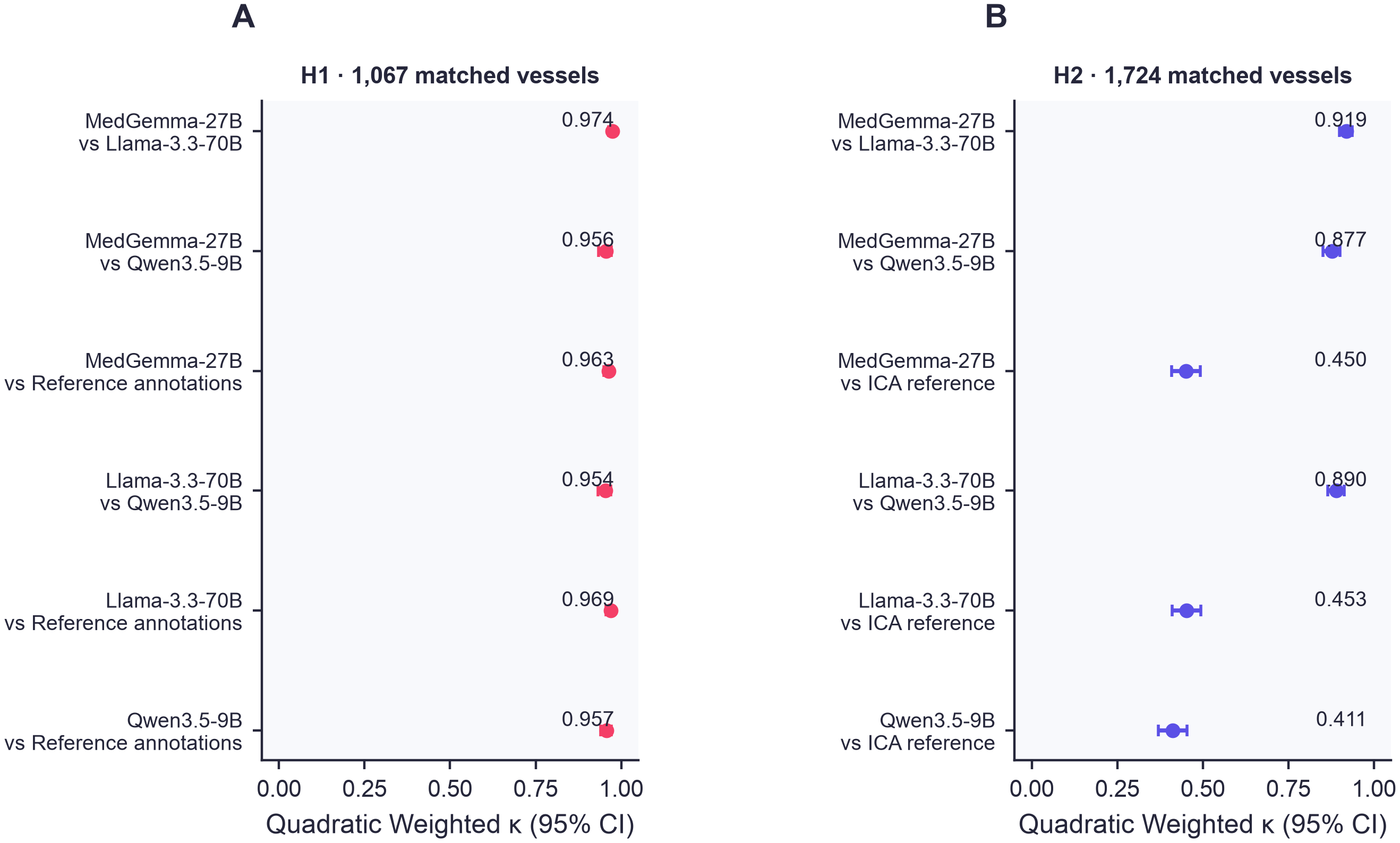}}
\end{figure}

\begin{figure}[t]
\floatconts
  {fig:roc-extraction}
  {\caption{\textbf{Discrimination of extraction failure.}
  Apparent ROC curves in P2 (1{,}077 vessels from 253 patients; 110 extraction
  errors) for Stenosis Margin alone, the CAD-phenotype baseline, and the baseline
  with Stenosis Margin added. The baseline includes LLM-extracted CCTA stenosis
  severity, calcium burden with a not-reported indicator, and vessel identity.
  Adding Stenosis Margin increased AUC from 0.805 to 0.862
  ($\Delta$AUC $+0.057$, 95\% CI $+0.029$ to $+0.087$; $P=0.001$).
  Confidence intervals and the paired AUC comparison use 2{,}000
  patient-cluster bootstrap resamples.}}
  {\includegraphics[width=0.85\linewidth]{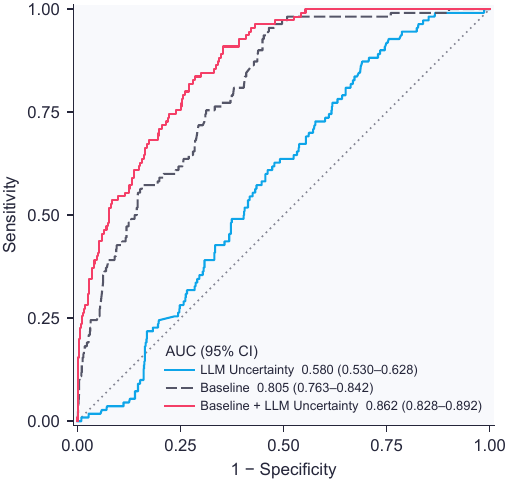}}
\end{figure}

\begin{figure}[t]
\floatconts
  {fig:roc-clinical}
  {\caption{\textbf{Discrimination of clinical discordance.}
  Apparent ROC curves in P3 (1{,}641 vessels from 729 patients; 476 clinically
  discordant vessels) for Stenosis Margin alone, the CAD-phenotype baseline,
  and the baseline with Stenosis Margin added. Clinical discordance is
  disagreement between LLM-extracted CCTA and ICA measurements on obstructive
  CAD status, defined as $\geq 50\%$ stenosis in the left main and
  $\geq 70\%$ in the other vessels. Baseline covariates are as in
  \figureref{fig:roc-extraction}. Adding Stenosis Margin changed AUC from
  0.618 to 0.625 ($\Delta$AUC $+0.007$, 95\% CI $-0.004$ to $+0.017$;
  $P=0.245$). Confidence intervals and the paired AUC comparison use
  2{,}000 patient-cluster bootstrap resamples.}}
  {\includegraphics[width=0.85\linewidth]{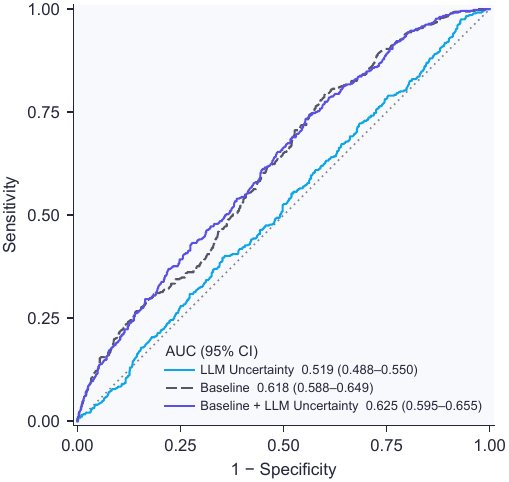}}
\end{figure}

\clearpage
\subsection{Supplementary tables}

\begin{table*}[t]
\floatconts
  {tab:llm-regex}
  {\caption{\textbf{LLMs versus regex against radiologist annotations.}
  Top: weighted $\kappa$ and regex-minus-LLM differences (95\% CIs); $P$ values use
  2{,}000 paired patient-bootstrap resamples with Holm correction across three comparisons.
  Bottom: exact agreement, number (\%) within each reference stenosis category.}}
  {\scriptsize\setlength{\tabcolsep}{2pt}\renewcommand{\arraystretch}{0.9}%
  \setlength{\aboverulesep}{1.5pt}\setlength{\belowrulesep}{1.5pt}%
  \begin{tabular*}{\linewidth}{@{\extracolsep{\fill}}lccc@{}}
    \toprule
    Method & Quadratic weighted $\kappa$ & Regex $-$ method, $\Delta\kappa$ & Holm-adjusted $P$ \\
    \midrule
    Regex & 0.946 (0.925--0.964) & --- & --- \\
    Llama 3.3 70B & 0.974 (0.966--0.981) & $-0.028$ ($-0.050$ to $-0.009$) & 0.003 \\
    MedGemma 27B & 0.963 (0.949--0.974) & $-0.017$ ($-0.040$ to 0.004) & 0.21 \\
    Qwen 3.5 9B & 0.957 (0.940--0.971) & $-0.011$ ($-0.034$ to 0.012) & 0.36 \\
    \bottomrule
  \end{tabular*}
  \par\vspace{3pt}
  \begin{tabular*}{\linewidth}{@{\extracolsep{\fill}}lcccccc@{}}
    \toprule
    Method & 0\% ($n=460$) & 1--24\% ($n=175$) & 25--49\% ($n=162$) & 50--69\% ($n=106$) & 70--99\% ($n=117$) & 100\% ($n=47$) \\
    \midrule
    Regex & 450 (97.8\%) & 144 (82.3\%) & 136 (84.0\%) & 92 (86.8\%) & 99 (84.6\%) & 45 (95.7\%) \\
    Llama 3.3 70B & 446 (97.0\%) & 157 (89.7\%) & 114 (70.4\%) & 92 (86.8\%) & 114 (97.4\%) & 36 (76.6\%) \\
    MedGemma 27B & 445 (96.7\%) & 142 (81.1\%) & 134 (82.7\%) & 87 (82.1\%) & 108 (92.3\%) & 35 (74.5\%) \\
    Qwen 3.5 9B & 450 (97.8\%) & 150 (85.7\%) & 123 (75.9\%) & 85 (80.2\%) & 106 (90.6\%) & 34 (72.3\%) \\
    \bottomrule
  \end{tabular*}}
\end{table*}

\begin{table}[t]
\floatconts
  {tab:s1}
  {\caption{Cohort flow. SOURCE rows count patients; the remaining rows count patient-vessel pairs.}}
  {\scriptsize\setlength{\tabcolsep}{5pt}%
  \begin{tabular}{llrr}
    \toprule
    Cohort & Step & Patients & Vessels \\
    \midrule
    SOURCE & index studies & 7,847 &  \\
    SOURCE & excluded prior ica & 255 &  \\
    SOURCE & excluded stent or cabg & 302 &  \\
    SOURCE & excluded no ica in window & 6,443 &  \\
    SOURCE & ica eligible & 847 &  \\
    P1 & Vessel rows built from the cohort export & 1,045 & 5,225 \\
    P2 & Vessels in reference-annotated studies & 270 & 1,350 \\
    P2 & With an LLM-extracted stenosis severity & 253 & 1,090 \\
    P2 & With an expert reference stenosis severity & 253 & 1,077 \\
    P2 & With all four uncertainty measurements (ANALYSED) & 253 & 1,077 \\
    P3 & Vessels in studies with an ICA within 0 to 90 days & 847 & 4,235 \\
    P3 & With an LLM-extracted stenosis severity & 789 & 3,370 \\
    P3 & With a quantified vessel-matched ICA severity & 729 & 1,641 \\
    P3 & With all four uncertainty measurements (ANALYSED) & 729 & 1,641 \\
    \bottomrule
  \end{tabular}}
\end{table}

\begin{table}[t]
\floatconts
  {tab:s2}
  {\caption{Odds ratios for every uncertainty measure, per one standard deviation more uncertain. Adjusted models add the CAD-phenotype baseline of CCTA stenosis severity, $\log_2(\mathrm{CACS}+1)$ with a not-reported indicator, and vessel. The two Findings measures are complete-case on a minority of vessels (Table~\ref{tab:s4}). Adjusted Findings Margin did not converge in P2 and is omitted; adjusted Findings Perplexity converged onto a degenerate interval and no P value is reported.}}
  {\scriptsize\setlength{\tabcolsep}{5pt}%
  \begin{tabular}{lllrrr}
    \toprule
    Analysis & Model & Measure & OR (95\% CI) & P & Holm P \\
    \midrule
    Extraction Failure & Univariable & Stenosis Margin & 1.10 (0.96--1.27) & 0.169 & 0.508 \\
    Extraction Failure & Univariable & Stenosis Perplexity & 1.21 (0.99--1.47) & 0.060 & 0.240 \\
    Extraction Failure & Univariable & Self-Confidence & 2.16 (1.77--2.64) & < 0.001 & < 0.001 \\
    Extraction Failure & Univariable & Self-Ambiguity & 1.75 (1.48--2.06) & < 0.001 & < 0.001 \\
    Extraction Failure & Univariable & Findings Margin & 1.12 (0.87--1.44) & 0.367 & 0.508 \\
    Extraction Failure & Univariable & Findings Perplexity & 1.14 (0.94--1.38) & 0.193 & 0.508 \\
    Extraction Failure & Univariable & Confidence Margin & 2.03 (1.71--2.42) & < 0.001 & < 0.001 \\
    Extraction Failure & Univariable & Confidence Perplexity & 1.28 (1.06--1.54) & 0.010 & 0.057 \\
    Extraction Failure & Univariable & Ambiguity Margin & 2.17 (1.78--2.65) & < 0.001 & < 0.001 \\
    Extraction Failure & Univariable & Ambiguity Perplexity & 1.18 (1.02--1.36) & 0.029 & 0.143 \\
    Extraction Failure & Adjusted & Stenosis Margin & 4.15 (2.57--6.69) & < 0.001 & < 0.001 \\
    Extraction Failure & Adjusted & Stenosis Perplexity & 1.29 (1.12--1.48) & < 0.001 & 0.002 \\
    Extraction Failure & Adjusted & Self-Confidence & 1.62 (1.25--2.09) & < 0.001 & 0.002 \\
    Extraction Failure & Adjusted & Self-Ambiguity & 1.44 (1.17--1.77) & < 0.001 & 0.003 \\
    Extraction Failure & Adjusted & Findings Perplexity & 1.53 (1.53--1.53) & --- & --- \\
    Extraction Failure & Adjusted & Confidence Margin & 0.92 (0.64--1.33) & 0.665 & 1.000 \\
    Extraction Failure & Adjusted & Confidence Perplexity & 0.98 (0.79--1.22) & 0.871 & 1.000 \\
    Extraction Failure & Adjusted & Ambiguity Margin & 1.66 (1.15--2.39) & 0.007 & 0.027 \\
    Extraction Failure & Adjusted & Ambiguity Perplexity & 1.17 (0.99--1.38) & 0.074 & 0.222 \\
    Clinical Discordance & Univariable & Stenosis Margin & 0.87 (0.78--0.97) & 0.009 & 0.073 \\
    Clinical Discordance & Univariable & Stenosis Perplexity & 0.98 (0.88--1.09) & 0.704 & 1.0 \\
    Clinical Discordance & Univariable & Self-Confidence & 1.18 (1.05--1.31) & 0.004 & 0.033 \\
    Clinical Discordance & Univariable & Self-Ambiguity & 1.22 (1.10--1.35) & < 0.001 & 0.002 \\
    Clinical Discordance & Univariable & Findings Margin & 1.09 (0.90--1.32) & 0.391 & 1.0 \\
    Clinical Discordance & Univariable & Findings Perplexity & 0.96 (0.82--1.12) & 0.608 & 1.0 \\
    Clinical Discordance & Univariable & Confidence Margin & 1.03 (0.93--1.15) & 0.552 & 1.0 \\
    Clinical Discordance & Univariable & Confidence Perplexity & 1.04 (0.94--1.16) & 0.463 & 1.0 \\
    Clinical Discordance & Univariable & Ambiguity Margin & 1.14 (1.02--1.28) & 0.021 & 0.15 \\
    Clinical Discordance & Univariable & Ambiguity Perplexity & 1.01 (0.90--1.12) & 0.880 & 1.0 \\
    Clinical Discordance & Adjusted & Stenosis Margin & 1.22 (1.04--1.43) & 0.017 & 0.118 \\
    Clinical Discordance & Adjusted & Stenosis Perplexity & 1.02 (0.91--1.15) & 0.720 & 1.0 \\
    Clinical Discordance & Adjusted & Self-Confidence & 1.16 (1.03--1.31) & 0.015 & 0.118 \\
    Clinical Discordance & Adjusted & Self-Ambiguity & 1.19 (1.07--1.33) & 0.002 & 0.018 \\
    Clinical Discordance & Adjusted & Findings Margin & 1.00 (0.79--1.28) & 0.967 & 1.0 \\
    Clinical Discordance & Adjusted & Findings Perplexity & 0.94 (0.81--1.08) & 0.372 & 1.0 \\
    Clinical Discordance & Adjusted & Confidence Margin & 1.04 (0.91--1.17) & 0.581 & 1.0 \\
    Clinical Discordance & Adjusted & Confidence Perplexity & 1.01 (0.91--1.13) & 0.823 & 1.0 \\
    Clinical Discordance & Adjusted & Ambiguity Margin & 1.25 (1.10--1.43) & < 0.001 & 0.009 \\
    Clinical Discordance & Adjusted & Ambiguity Perplexity & 1.02 (0.91--1.14) & 0.727 & 1.0 \\
    \bottomrule
  \end{tabular}}
\end{table}

\begin{table}[t]
\floatconts
  {tab:s3}
  {\caption{Discrimination and calibration for every model in both analyses. Intervals and P values come from 2{,}000 patient-cluster bootstrap replicates; optimism correction uses the Efron--Gong procedure over 400 replicates. The Findings rows are fitted complete-case and are not comparable with the rest.}}
  {\scriptsize\setlength{\tabcolsep}{3pt}%
  \begin{tabular}{lrrrrr}
    \toprule
    Model & \hd{AUC\\(95\% CI)} & \hd{$\Delta$AUC vs baseline\\(95\% CI)} & \hd{Bootstrap\\P} & \hd{Brier\\score} & \hd{Optimism-\\corrected AUC} \\
    \midrule
    \multicolumn{6}{l}{\emph{P2 --- Extraction Failure}} \\
    Baseline & 0.805 (0.763--0.842) & --- & --- & 0.0788 & 0.783 \\
    Stenosis Margin alone & 0.580 (0.530--0.628) & --- & --- & 0.0917 & --- \\
    Baseline + Stenosis Margin & 0.862 (0.828--0.892) & +0.0571 (+0.0290, +0.0872) & 0.001 & 0.0702 & 0.845 \\
    Stenosis Perplexity alone & 0.784 (0.738--0.824) & --- & --- & 0.0914 & --- \\
    Baseline + Stenosis Perplexity & 0.819 (0.779--0.854) & +0.0139 (+0.0028, +0.0268) & 0.014 & 0.0771 & 0.799 \\
    Self-Confidence alone & 0.727 (0.682--0.769) & --- & --- & 0.0868 & --- \\
    Baseline + Self-Confidence & 0.825 (0.785--0.860) & +0.0200 (+0.0047, +0.0361) & 0.015 & 0.0771 & 0.803 \\
    Self-Ambiguity alone & 0.729 (0.682--0.775) & --- & --- & 0.0884 & --- \\
    Baseline + Self-Ambiguity & 0.825 (0.785--0.862) & +0.0206 (+0.0058, +0.0373) & 0.007 & 0.0772 & 0.804 \\
    Findings Perplexity alone & 0.541 (0.444--0.647) & --- & --- & 0.0868 & --- \\
    Baseline + Findings Perplexity & 0.826 (0.770--0.881) & +0.0339 ($-$0.0352, +0.0787) & 0.448 & 0.073 & 0.781 \\
    Confidence Margin alone & 0.688 (0.646--0.730) & --- & --- & 0.089 & --- \\
    Baseline + Confidence Margin & 0.804 (0.762--0.841) & $-$0.0004 ($-$0.0042, +0.0030) & 0.836 & 0.0787 & 0.78 \\
    Confidence Perplexity alone & 0.690 (0.644--0.736) & --- & --- & 0.0909 & --- \\
    Baseline + Confidence Perplexity & 0.804 (0.762--0.841) & $-$0.0007 ($-$0.0036, +0.0015) & 0.654 & 0.0787 & 0.779 \\
    Ambiguity Margin alone & 0.716 (0.667--0.761) & --- & --- & 0.0882 & --- \\
    Baseline + Ambiguity Margin & 0.820 (0.781--0.855) & +0.0154 (+0.0005, +0.0312) & 0.045 & 0.0781 & 0.798 \\
    Ambiguity Perplexity alone & 0.676 (0.624--0.723) & --- & --- & 0.0911 & --- \\
    Baseline + Ambiguity Perplexity & 0.810 (0.769--0.846) & +0.0054 ($-$0.0023, +0.0134) & 0.181 & 0.0784 & 0.787 \\
    \addlinespace
    \multicolumn{6}{l}{\emph{P3 --- Clinical Discordance}} \\
    Baseline & 0.618 (0.588--0.649) & --- & --- & 0.1968 & 0.602 \\
    Stenosis Margin alone & 0.519 (0.488--0.550) & --- & --- & 0.2053 & --- \\
    Baseline + Stenosis Margin & 0.625 (0.595--0.655) & +0.0067 ($-$0.0042, +0.0175) & 0.245 & 0.1961 & 0.608 \\
    Stenosis Perplexity alone & 0.479 (0.450--0.510) & --- & --- & 0.2059 & --- \\
    Baseline + Stenosis Perplexity & 0.618 (0.587--0.648) & $-$0.0004 ($-$0.0027, +0.0020) & 0.69 & 0.1968 & 0.6 \\
    Self-Confidence alone & 0.538 (0.508--0.567) & --- & --- & 0.2047 & --- \\
    Baseline + Self-Confidence & 0.625 (0.594--0.657) & +0.0067 ($-$0.0035, +0.0173) & 0.212 & 0.196 & 0.608 \\
    Self-Ambiguity alone & 0.543 (0.513--0.574) & --- & --- & 0.204 & --- \\
    Baseline + Self-Ambiguity & 0.630 (0.601--0.661) & +0.0119 (+0.0002, +0.0239) & 0.049 & 0.1955 & 0.614 \\
    Findings Margin alone & 0.499 (0.442--0.555) & --- & --- & 0.1971 & --- \\
    Baseline + Findings Margin & 0.657 (0.601--0.709) & +0.0285 ($-$0.0099, +0.0868) & 0.123 & 0.1837 & 0.615 \\
    Findings Perplexity alone & 0.454 (0.393--0.515) & --- & --- & 0.1973 & --- \\
    Baseline + Findings Perplexity & 0.659 (0.605--0.711) & +0.0305 ($-$0.0085, +0.0885) & 0.103 & 0.1835 & 0.618 \\
    Confidence Margin alone & 0.499 (0.468--0.531) & --- & --- & 0.2059 & --- \\
    Baseline + Confidence Margin & 0.618 (0.588--0.649) & $-$0.0001 ($-$0.0036, +0.0035) & 0.935 & 0.1968 & 0.6 \\
    Confidence Perplexity alone & 0.528 (0.497--0.559) & --- & --- & 0.2059 & --- \\
    Baseline + Confidence Perplexity & 0.618 (0.588--0.648) & $-$0.0005 ($-$0.0019, +0.0010) & 0.548 & 0.1968 & 0.599 \\
    Ambiguity Margin alone & 0.536 (0.503--0.565) & --- & --- & 0.2052 & --- \\
    Baseline + Ambiguity Margin & 0.633 (0.603--0.662) & +0.0145 ($-$0.0009, +0.0288) & 0.064 & 0.1952 & 0.616 \\
    Ambiguity Perplexity alone & 0.525 (0.493--0.556) & --- & --- & 0.2059 & --- \\
    Baseline + Ambiguity Perplexity & 0.618 (0.588--0.649) & $-$0.0000 ($-$0.0024, +0.0024) & 0.995 & 0.1968 & 0.6 \\
    \bottomrule
  \end{tabular}}
\end{table}

\begin{table}[t]
\floatconts
  {tab:s4}
  {\caption{Availability of each uncertainty measure. The two Findings measures are complete-case; every other measure is present for all vessels in its population.}}
  {\scriptsize\setlength{\tabcolsep}{5pt}%
  \begin{tabular}{llrrl}
    \toprule
    Analysis & Measure & Vessels & Present (\%) & Sparsest vessel \\
    \midrule
    Extraction Failure & Stenosis Margin & 1,077 & 100.0 & --- \\
    Extraction Failure & Stenosis Perplexity & 1,077 & 100.0 & --- \\
    Extraction Failure & Self-Confidence & 1,077 & 100.0 & --- \\
    Extraction Failure & Self-Ambiguity & 1,077 & 100.0 & --- \\
    Extraction Failure & Findings Margin & 1,077 & 46.0 & LAD \\
    Extraction Failure & Findings Perplexity & 1,077 & 46.0 & LAD \\
    Extraction Failure & Confidence Margin & 1,077 & 100.0 & --- \\
    Extraction Failure & Confidence Perplexity & 1,077 & 100.0 & --- \\
    Extraction Failure & Ambiguity Margin & 1,077 & 100.0 & --- \\
    Extraction Failure & Ambiguity Perplexity & 1,077 & 100.0 & --- \\
    Clinical Discordance & Stenosis Margin & 1,641 & 100.0 & --- \\
    Clinical Discordance & Stenosis Perplexity & 1,641 & 100.0 & --- \\
    Clinical Discordance & Self-Confidence & 1,641 & 100.0 & --- \\
    Clinical Discordance & Self-Ambiguity & 1,641 & 100.0 & --- \\
    Clinical Discordance & Findings Margin & 1,641 & 32.0 & RI \\
    Clinical Discordance & Findings Perplexity & 1,641 & 32.0 & RI \\
    Clinical Discordance & Confidence Margin & 1,641 & 100.0 & --- \\
    Clinical Discordance & Confidence Perplexity & 1,641 & 100.0 & --- \\
    Clinical Discordance & Ambiguity Margin & 1,641 & 100.0 & --- \\
    Clinical Discordance & Ambiguity Perplexity & 1,641 & 100.0 & --- \\
    \bottomrule
  \end{tabular}}
\end{table}

\begin{table}[t]
\floatconts
  {tab:s5}
  {\caption{The twenty most strongly agreeing pairs among the ten uncertainty measures, in the Extraction Failure population. Pairwise comparisons within the margin and perplexity families are plotted in Figure~\ref{fig:s5}.}}
  {\scriptsize\setlength{\tabcolsep}{5pt}%
  \begin{tabular}{llrr}
    \toprule
    Measure A & Measure B & Spearman $\rho$ & Pearson r \\
    \midrule
    Self-Confidence & Self-Ambiguity & 0.99 & 0.92 \\
    Confidence Margin & Confidence Perplexity & 0.88 & 0.53 \\
    Ambiguity Margin & Ambiguity Perplexity & 0.86 & 0.37 \\
    Self-Ambiguity & Ambiguity Perplexity & 0.84 & 0.34 \\
    Confidence Margin & Ambiguity Margin & 0.82 & 0.83 \\
    Self-Confidence & Ambiguity Perplexity & 0.82 & 0.26 \\
    Self-Ambiguity & Ambiguity Margin & 0.8 & 0.73 \\
    Self-Confidence & Ambiguity Margin & 0.79 & 0.78 \\
    Confidence Margin & Ambiguity Perplexity & 0.77 & 0.15 \\
    Confidence Perplexity & Ambiguity Margin & 0.76 & 0.39 \\
    Self-Confidence & Confidence Margin & 0.74 & 0.71 \\
    Self-Ambiguity & Confidence Margin & 0.72 & 0.58 \\
    Confidence Perplexity & Ambiguity Perplexity & 0.7 & 0.11 \\
    Self-Ambiguity & Confidence Perplexity & 0.64 & 0.28 \\
    Self-Confidence & Confidence Perplexity & 0.64 & 0.31 \\
    Stenosis Perplexity & Self-Confidence & 0.63 & 0.1 \\
    Stenosis Perplexity & Ambiguity Margin & 0.63 & 0.14 \\
    Stenosis Perplexity & Self-Ambiguity & 0.62 & 0.08 \\
    Stenosis Perplexity & Confidence Margin & 0.58 & 0.09 \\
    Stenosis Perplexity & Ambiguity Perplexity & 0.57 & 0.11 \\
    \bottomrule
  \end{tabular}}
\end{table}

\begin{table}[t]
\floatconts
  {tab:llm-reference-summary}
  {\caption{\textbf{Agreement against radiologist annotations and ICA.}
  Quadratic weighted $\kappa$ (95\% CI). LLM estimates are from the matched-model
  analysis in \figureref{fig:llm-agreement}; regex estimates are those reported during
  review. Paired comparisons against regex appear in \tableref{tab:llm-regex,tab:regex-ica}.}}
  {\scriptsize\setlength{\tabcolsep}{2pt}\renewcommand{\arraystretch}{0.9}%
  \setlength{\aboverulesep}{1.5pt}\setlength{\belowrulesep}{1.5pt}%
  \begin{tabular*}{\linewidth}{@{\extracolsep{\fill}}lcc@{}}
    \toprule
    Method & Assigned radiologist & ICA reference \\
    \midrule
    Llama 3.3 70B & 0.969 (0.956--0.980) & 0.453 (0.410--0.494) \\
    MedGemma 27B & 0.963 (0.949--0.974) & 0.450 (0.409--0.493) \\
    Qwen 3.5 9B & 0.957 (0.940--0.971) & 0.411 (0.370--0.453) \\
    Regex & 0.946 (0.925--0.964) & 0.335 (0.294--0.378) \\
    \bottomrule
  \end{tabular*}}
\end{table}

\begin{table}[t]
\floatconts
  {tab:regex-ica}
  {\caption{\textbf{LLMs versus regex against ICA.}
  Top: weighted $\kappa$ and regex-minus-LLM differences (95\% CIs), as reported during
  review; $P$ values use 2{,}000 paired patient-bootstrap resamples with Holm correction
  across three comparisons. Bottom: exact agreement, number (\%) within each ICA
  stenosis category. The empty ICA 0\% category is omitted.}}
  {\scriptsize\setlength{\tabcolsep}{2pt}\renewcommand{\arraystretch}{0.9}%
  \setlength{\aboverulesep}{1.5pt}\setlength{\belowrulesep}{1.5pt}%
  \begin{tabular*}{\linewidth}{@{\extracolsep{\fill}}lccc@{}}
    \toprule
    Method & Quadratic weighted $\kappa$ & Regex $-$ method, $\Delta\kappa$ & Holm-adjusted $P$ \\
    \midrule
    Regex & 0.335 (0.294--0.378) & --- & --- \\
    Llama 3.3 70B & 0.454 (0.412--0.494) & $-0.119$ ($-0.151$ to $-0.087$) & 0.003 \\
    MedGemma 27B & 0.450 (0.409--0.493) & $-0.115$ ($-0.149$ to $-0.080$) & 0.003 \\
    Qwen 3.5 9B & 0.411 (0.370--0.453) & $-0.076$ ($-0.111$ to $-0.040$) & 0.003 \\
    \bottomrule
  \end{tabular*}
  \par\vspace{3pt}
  \begin{tabular*}{\linewidth}{@{\extracolsep{\fill}}lccccc@{}}
    \toprule
    Method & 1--24\% ($n=22$) & 25--49\% ($n=483$) & 50--69\% ($n=358$) & 70--99\% ($n=780$) & 100\% ($n=81$) \\
    \midrule
    Regex & 5 (22.7\%) & 125 (25.9\%) & 117 (32.7\%) & 317 (40.6\%) & 52 (64.2\%) \\
    Llama 3.3 70B & 5 (22.7\%) & 110 (22.8\%) & 127 (35.5\%) & 461 (59.1\%) & 43 (53.1\%) \\
    MedGemma 27B & 2 (9.1\%) & 137 (28.4\%) & 149 (41.6\%) & 437 (56.0\%) & 40 (49.4\%) \\
    Qwen 3.5 9B & 4 (18.2\%) & 123 (25.5\%) & 124 (34.6\%) & 409 (52.4\%) & 42 (51.9\%) \\
    \bottomrule
  \end{tabular*}}
\end{table}

\end{document}